\pdfoutput=1

\documentclass[11pt]{article}

\usepackage{ACL2023}
\usepackage{times}
\usepackage{latexsym}
\usepackage{pifont}
\usepackage{enumitem}
\usepackage{amsmath}
\usepackage{rotating}
\usepackage{listings}
\usepackage{subfigure}
\usepackage{float}
\usepackage{enumitem}
\setlist[itemize]{align=parleft,left=0.0em}

\usepackage[T1]{fontenc}
\usepackage{xspace}

\usepackage[utf8]{inputenc}
\usepackage{booktabs}
\usepackage{microtype}
\usepackage{multirow}
\usepackage{rotating}
\newcommand{\datasetName}{{\sc NTSEBench}\xspace}
\usepackage{inconsolata}
\usepackage{tcolorbox}
\usepackage{hyperref}

\title{
\textbf{\datasetName}: Cognitive Reasoning Benchmark for Vision Language Models}

\author{
Pranshu Pandya\textsuperscript{\rm 1†},
Vatsal Gupta\textsuperscript{\rm 1†},
Agney S Talwarr\textsuperscript{\rm 1} \\ 
\textbf{Tushar Kataria}\textsuperscript{\rm 2},
\textbf{Dan Roth}\textsuperscript{\rm 3},
\textbf{Vivek Gupta}\textsuperscript{\rm 4}\thanks{~~Corresponding Author, †Equal Contribution}~\\
\textsuperscript{\rm 1}IIT Guwahati,
\textsuperscript{\rm 2}University of Utah \\
\textsuperscript{\rm 3}University of Pennsylvania,
\textsuperscript{\rm 4}Arizona State University\\
\small \{p.pandya,g.vatsal,t.agney\}@iitg.ac.in, tkataria@cs.utah.edu, danroth@seas.upenn.edu, vgupt140@asu.edu 
}

\begin{document}
\maketitle
\begin{abstract}
Cognitive textual and visual reasoning tasks, including puzzles, series, and analogies, demand the ability to quickly reason, decipher, and evaluate patterns both textually and spatially. Due to extensive training on vast amounts of human-curated data, large language models (LLMs) and vision language models (VLMs) excel in common-sense reasoning tasks, but still struggle with more complex reasoning that demands deeper cognitive understanding. We introduce \datasetName, a new dataset designed to evaluate cognitive multimodal reasoning and problem-solving skills of large models. The dataset contains 2,728 multiple-choice questions, accompanied by a total of 4,642 images, spanning 26 categories. These questions are drawn from the nationwide NTSE examination in India and feature a mix of visual and textual general aptitude challenges, designed to assess intelligence and critical thinking skills beyond mere rote learning. We establish baselines on the dataset using state-of-the-art LLMs and VLMs. To facilitate a comparison between open-source and propriety models, we propose four distinct modeling strategies to handle different modalities—text and images—in the dataset instances.
\end{abstract}
\section{Introduction}
Aptitude and reasoning tests have been essential for assessing intelligence and are considered strong indicators of problem-solving ability and abstract reasoning skills \cite{stern1914psychological}. Recent advancements in large language models (LLMs) have demonstrated their strong performance on IQ test questions, achieving high scores across many languages \cite{king2023administration}. These results indicate that LLMs are advancing toward pseudo human-like intelligence, particularly in text and language tasks. 

\begin{figure}
    \centering
    \includegraphics[scale=0.47]{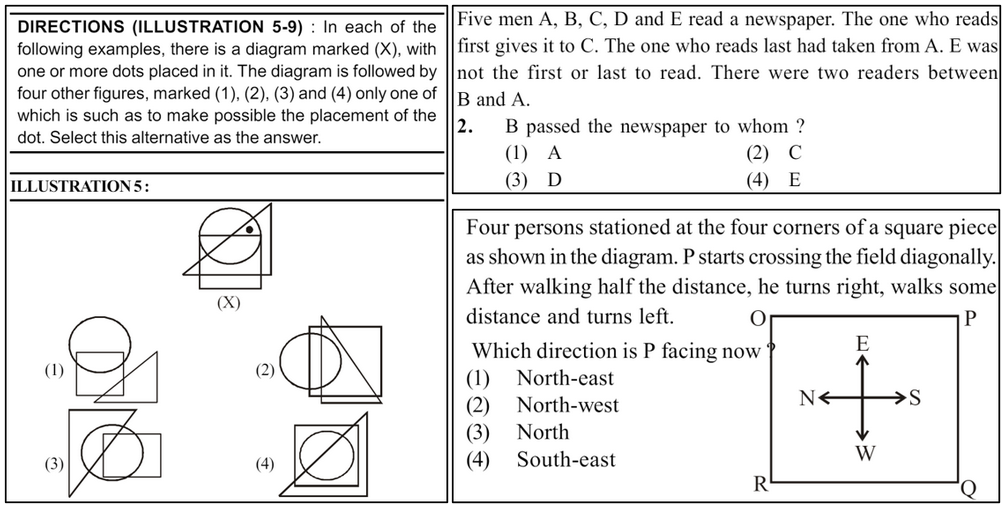}
    \caption{\textbf{\datasetName Examples}: Three samples of textual, direction, and spatial reasoning questions from the proposed dataset. Solutions to these questions are not included here but are provided in the dataset.}
    \label{fig:1}

\end{figure}

The capabilities of LLM models rivals humans on various tasks—question answering (QA), sentiment classification, text generation, visual QA, coding challenges, and mathematical reasoning, to name a few \cite{srivastava2022beyond,bubeck2023sparks}. LLMs and VLMs (also known as Multimodal Large Language Models, MLLMs) have hence become the default benchmark for many zero or few-shot text and vision-based tasks \cite{brown2020language,wei2022chain}.  
Training on vast datasets from diverse domains has enabled LLMs to achieve human-level performance in SAT, GRE, and AP exams, and on platforms such as LeetCode \cite{achiam2023gpt,touvron2023llama,dubey2024llama}.

The results from \citet{king2023administration} indicate that LLMs excel in tasks involving textual reasoning such as comprehension, analogies, and identifying opposites but struggle with other types of questions. 
Multimodal Large Language models have demonstrated remarkable performance in many tests of human intelligence, but they still fall short of human baselines in tasks that require critical and logical thinking, such as commonsense-, numerical- and scientific-reasoning, puzzles, and analogies.
Most existing visual and multimodal reasoning datasets are domain-specific, concentrating on fields such as science, engineering, and medicine \cite{yue2024mmmu,zhang2024m3exam,sun2024scieval}. 
These datasets primarily focus on tasks related to concrete scenarios or specific domains, often requiring domain knowledge and rote learning for high performance. However, they do not adequately assess intelligence as a function of cognitive/critical reasoning skills such as spatial recognition, visual puzzle solving, abstract reasoning, or pattern recognition.
In this study, we aim to address this research gap by introducing a novel benchmark dataset, \datasetName, specifically created to evaluate the complex visual, textual, and multimodal cognitive reasoning capabilities of large deep learning models. Examples of questions from the proposed dataset are shown in Figure \ref{fig:1}.

\datasetName is dedicated to establishing a benchmark for testing capabilities that do not rely on domain-specific knowledge or rote learning. Its primary contribution lies in evaluating the innate problem-solving skills inherent in human cognitive development and isolating where models are lacking by presenting well-categorized data. 
It comprises questions sourced from the Nationwide Talent Search Examination (NTSE) conducted in India.
These questions can be presented in text format, visual format, or both (multimodal).
We evaluate the performance of recent LLMs and VLMs, including both proprietary  \cite{achiam2023gpt,reid2024gemini} and open-source models \cite{achiam2023gpt,touvron2023llama,jiang2024mixtral,wang2023cogvlm,bai2023qwen, li2024llava, lu2024ovis} on our dataset and perform in-depth error analysis of LLM responses to pinpoint areas of weakness and evaluate their overall robustness.  Our work makes the following contributions:
\vspace{-0.5em}
\begin{itemize}
\setlength\itemsep{-0.25em}
    \item \datasetName, a dataset to evaluate complex textual, visual, and multimodal cognitive reasoning capabilities with 2,728 questions in 26 problem categories.
    \item Establish baselines using a diverse range of state-of-the-art LLMs and VLMs on the proposed dataset, incorporating both open-source and proprietary models.
    \item Assess performance using various modeling strategies to effectively handle multimodal (multi-image inputs as well) inputs for reasoning-based questions.
\end{itemize}
Code and dataset for the experiments with \datasetName are available at \href{https://ntsebench.github.io/}{https://ntsebench.github.io/}.

\section{\datasetName Benchmark}
\label{sec:ntse_benchmark}
\quad \textbf{NTSE Exam.} The National Talent Search Examination (NTSE), administered by the National Council of Educational Research and Training (NCERT) in India since 1963, is a nationwide exam for secondary-grade students. The exam consists of two sections designed to assess a wide range of analytical skills: the Mental Ability Test (MAT) and the Scholastic Aptitude Test (SAT). The MAT section evaluates students' general aptitude, critical thinking, logical and spatial reasoning, and analytical problem-solving skills (for both textual and visual problems). In contrast, the SAT assesses their domain-specific knowledge in science and mathematics. All questions in the NTSE are multiple-choice (MCQ) with one correct option. Questions and options can be text or image or a combination of both, i.e., multimodal. 
Our aim is to create a dataset focused to test cognitive reasoning abilities (or MAT-type questions).

\label{cog-reasn}
\textbf{Cognitive Reasoning.} Cognitive understanding in the context of \datasetName refers to the ability to process information, recognize patterns, draw inferences, and solve problems using critical, logical, and analytical reasoning. This aligns with fundamental concepts in cognitive science, such as problem-solving, pattern recognition, and inferential reasoning \cite{wang2010cognitive}. It encompasses advanced reasoning skills typically found in a small subset of the population, generally individuals with very high IQs. To emphasize this distinction, we use the term \textit{cognitive reasoning} for our dataset, differentiating it from common sense reasoning tasks \cite{sakaguchi2021winogrande,talmor-etal-2019-commonsenseqa}. \datasetName assesses these reasoning abilities through diverse question categories, targeting a different cognitive dimension:
\begin{itemize}
\setlength\itemsep{-0.25em}
    \item  \textbf{Pattern Recognition}: Categories such as Series (Numerical, Alphabetical, Alphanumeric), Missing Character, Non-verbal Series, and Dot Problem test the ability to identify and extend patterns, which is crucial for understanding sequences and predicting outcomes.
    \item \textbf{Logical Deduction}: Blood Relation, Syllogisms, Statement and Conclusions, and Data Sufficiency categories focus on making inferences and drawing conclusions based on the given information, reflecting the core of logical reasoning.
    \item \textbf{Spatial Reasoning}: Direction Sense, Cube and Dice, Paper Folding and Cutting, and Embedded Figure assess the ability to visualize and manipulate objects in space, which is essential for understanding spatial relationships.
    \item \textbf{Relational Reasoning}: Analogy and Non-verbal Analogy categories evaluate the understanding of relationships between items and the ability to transfer this understanding to new contexts, a key component of relational reasoning.
    \item \textbf{Quantitative Analysis}: Number and Ranking, Mathematical Operations, Time and Clock, and Figure Partition test numerical problem-solving skills and the ability to manage quantitative data.
    \item \textbf{Classification and Categorization}: Classification/Odd One Out, Non-verbal Classification/Odd One Out categories measure the ability to group items based on shared attributes and identify outliers, highlighting skills in distinguishing unique characteristics and grouping.
    \item \textbf{Contextual Interpretation}:
    Looking for specific details, instructions, or constraints that are critical to understanding and solving the problem.
    \item \textbf{Verbal Reasoning}:
    Understanding semantic relationships, word meanings, and analogies.
\end{itemize}
Unlike other benchmarks that focus on specific academic domains or individual cognitive dimensions as stated above, \datasetName emphasizes a wide range of cognitive skills, offering a more comprehensive assessment of reasoning abilities. 

\subsection{Dataset Sources}\label{section:dataset_sources}

We created the dataset using past NTSE papers and solutions from Resonance.
Additionally, we used NTSE preparation materials, such as a reference book titled \textit{A Modern Approach to Verbal and Non-Verbal Reasoning}, 
which includes additional logical reasoning problems. We also incorporated content from another book titled \textit{Study Guide for NTSE} 
to construct our dataset. The question extraction process is detailed in Appendix \ref{ref:appendix:ExtractionPipeline}. The example questions of the released test dataset in Figure \ref{fig:1} and Appendix \ref{Appendix:Examples}.

\textbf{Problem Categories}: \datasetName encompasses several problem categories, each designed to test a distinct set of skills. Questions from these categories frequently appear in NTSE exams year after year. A detailed description of each category is done in Appendix Table~\ref{tab:category_desc} and examples are shown in Appendix section ~\ref{Appendix:Examples}.
More dataset related information is present in Appendix section \ref{ref:datasetDetails}.

\begin{table}[h!]
\scriptsize
\centering
\setlength{\tabcolsep}{1.8pt}
\begin{tabular}{lc|lc}
\hline
\multicolumn{2}{c|}{\bf Text Only} & \multicolumn{2}{c}{\bf Vision + Text} \\
\hline
\textbf{Categories} & \textbf{\# Samples} & \textbf{Categories} & \textbf{\# Samples}\\
\hline
Series &  256 & Non-Verbal Series &  95  \\
Alphabet Test &  94 & Missing Character &  127 \\
Odd one out &  170 & Embedded Figure &  96 \\
Analogy &  151 & Non-Verbal odd one out &  70 \\
Coding-Decoding &  149 & Non-Verbal Analogy &  100  \\
Number and Ranking &  139 & Paper Folding \& Cutting &  96 \\
Blood Relation &  126 & Incomplete Figure &  94 \\
Mathematical Operations &  99 & Figure Partition &  71 \\
Puzzle Test &  95 & Cube and Dice &  89 \\
Syllogisms &  44 & Dot problem &  23 \\
Statement \& Conclusions &  104 & Direction Sense &  96 \\
Data Sufficiency &  90  & Time and Clock &  51 \\
&   & Mirror, Water and Images &  92 \\
&   & Venn diagrams &  111 \\
\vspace{-2em}
\end{tabular}

\caption{\textbf{NTSEBench categories count}: Problem categories with different input modality types and number of samples for each.}\label{tab:dataset_cat_count}
\end{table}
Table \ref{tab:dataset_cat_count} shows a skewed distribution across various question categories. Notably, textual categories such as the Alphabet Test (ALP) and Mathematical Operations (MTO) contain 94 and 99 examples, respectively. In contrast, many vision-based categories are more challenging and typically include between 80 and 100 examples. For instance, the Non-Verbal Analogy (NVA) category, one of the most difficult, comprises 100 examples. Although this skewed distribution could impact model performance, exploring its effects is beyond the scope of this manuscript and is left for future work.

\paragraph{Modality Variations.} Since \datasetName has multimodal questions, options, and solutions, we have results in eight combinations of modality types that can occur for question-options-solution triplet. Table \ref{tab:modality_count} shows the count of each triplet option. \datasetName has 1199 textual questions and the remaining 1529 are multimodal questions. 

\begin{table}[!htb]
\scriptsize
\centering
\begin{tabular}{cccc}
\hline
\textbf{Question} & \textbf{Options} & \textbf{Solutions} & \textbf{\# Samples} \\
\hline
\ding{55} & \ding{55} & \ding{55} & 1199 \\
\ding{55} & \ding{55} & \ding{51} & 381 \\
\ding{55} & \ding{51} & \ding{55} & 70 \\
\ding{55} & \ding{51} & \ding{51} & 18 \\
\ding{51} & \ding{55} & \ding{55} & 330 \\
\ding{51} & \ding{55} & \ding{51} & 126 \\
\ding{51} & \ding{51} & \ding{55} & 403 \\
\ding{51} & \ding{51} & \ding{51} & 201 \\
\hline
\end{tabular}
\caption{\textbf{\datasetName Modality Variations Question Count}: Tick(\ding{51}) mark indicates whether question, option or solution contains image. }\label{tab:modality_count}
\end{table}
\subsection{The Global Relevance of the NTSE Exam for AI}
\label{sec:relavance}
The NTSE exam, despite being conducted in India, holds significant relevance for the global AI community due to its unique focus on cognitive reasoning abilities rather than domain-specific knowledge. The NTSE's diverse question categories as described in Table ~\ref{tab:dataset_cat_count}( and Appendix table ~\ref{tab:category_desc}), assess various cognitive dimensions, offering a robust framework for evaluating AI models' capacity to process information, recognize patterns, and solve problems across different domains. This emphasis on cognitive dimensions aligns with the pursuit of Artificial General Intelligence (AGI), making the NTSE exam's insights and challenges applicable to AI research and development on a global scale.

\section{Models: LLMs and VLMs}

\textbf{Problem Formulation.} Consider a single ($i^{th}$) instance in the dataset is represented by {\sc D}$_i$=({\sc Q}$^J_i$,{\sc O}$^J_i$,{\sc S}$^J_i$), where {\sc Q} represents the questions, {\sc O} represents the options of the MCQ, and {\sc S} represents the solution to the question. $J \in (T,I)$ represents the modality type, which can be either text($T$) or image($I$).

\paragraph{Modeling Strategies.}
Evaluating the reasoning abilities of large language models (LLMs) with text-based questions is straightforward. For vision-language models (VLMs), reasoning with vision-text questions is generally not straightforward. 
Some API access model models, such as GPT-4o \cite{achiam2023gpt} and Gemini \cite{reid2024gemini}, support multi-image inputs, but many others do not (open-source models like LLaVA-OneVision \cite{li2024llava} and Ovis \cite{lu2024ovis} are emerging with this capability). To address these task-specific and input-related dependencies, we propose four strategies to fairly evaluate the reasoning abilities of both open-source and proprietary models.
\vspace{-0.5em}
\begin{itemize}
    \item \textbf{\textit{Standard QA}.} For instances where question type($J$) for questions($Q$), options($O$) and solutions($S$) is text($T$), we use a standard text-based QA model such as GPT3.5-Turbo or Llama3-70b \cite{llama3modelcard} or Mixtral8x7b \cite{jiang2024mixtral}. 
    
\item \textbf{\textit{Image-Only}.} We propose a modeling approach where questions and all the options are presented to the model as a single image. This image consolidates all relevant textual and visual content exactly as it appears in the examination paper, effectively capturing the entire question, including both textual and visuals. This strategy utilizes the OCR capabilities of VLM models to interpret and analyze the content, enabling them to process both text and visual elements within the same input \cite{shi2023exploring,fujitake2024dtrocr,zhao2023clip4str}. The key advantage of this approach is its applicability across all modality types.
    
\item \textbf{\textit{Interleaved model}.} In this approach, we integrate text with multiple images to create an interwoven context. This method involves placing related textual and visual elements in proximity, enhancing the model's ability to draw connections.

\item \textbf{\textit{Standard VQA}.} Open-source models typically lack the capability to integrate text and images within a single prompt. To enable fair comparisons, we propose an alternative modeling strategy where the question and option images are combined into a single composite image, labeled as Figure 1, Figure 2, etc. This composite image is accompanied by a structured textual prompt that describes different parts of the image, directing the model's attention to relevant visual details. The composite image and prompt are then used to evaluate the model's performance, testing its ability to interpret and respond to questions based on the integrated visual and textual information.
\end{itemize}
Example inputs for each of the above modeling strategy proposed are shown in Appendix Figure \ref{fig:system_prompt}.

\textbf{Prompting Strategies.}
We mainly employed two main prompting strategies for setting up all the baselines on the proposed dataset: 
(A) \textbf{Zero Shot COT}: The model is presented with a prompt that includes a question and a set of answer options. It is tasked with selecting the correct option and providing an explanation that justifies its choice. (B) \textbf{Few Shot COT}: In few-shot chain-of-thought (COT) prompting, a set of $N$ exemplar triplets—each containing a question, options, and a solution ($D_i$)—is included in the prompt before presenting the test question. The number of exemplars $N$ is selected based on the token limit supported by the model.

 \textbf{Implementation Details.}
We evaluate \datasetName using multiple open-source and  proprietary LLMs \cite{achiam2023gpt,touvron2023llama,team2023gemini} and VLMs \cite{reid2024gemini,bai2023qwen,dong2024internlm,wang2023cogvlm}. We used a low temperature setting to promote reproducibility. Details on models used and their hyperparameters can be found in Appendix ~\ref{ref:appx:model_hyper}.

In the few-shot settings, the questions were sorted by solution length, and then an annotator picked the three examples with the most comprehensive responses.

 \textbf{Answer Extraction and Evaluation.} The answer extraction module uses a rule-based approach to identify the correct option from the response \cite{gupta2023multi}. Responses where the answers cannot be extracted by module are evaluated by humans. We used percentage accuracy as our \textbf{metric for evaluation} across all the models. If the model chooses the right option in the MCQ question, label is set as {\sc True}, else label is {\sc False}. 
 
\textbf{Option Bias Ablation.} We conducted option shuffling experiments to assess whether model performance is influenced by the position of the correct option and to detect potential bias. Four dataset variations were created, placing the correct answer in positions 1, 2, 3, and 4, respectively. Each experiment was run in three rounds, and results were averaged to ensure robustness.

\section{Results and Analysis}

Our experiments answer the following questions: 
\begin{itemize}
    \item How well do LLMs and VLMs perform on advanced textual, visual, and multimodal reasoning questions? How challenging is cognitive reasoning for the current state-of-the-art models? 
    \item Are proprietary models superior to open-source models, and if so then by what margin? Are there specific categories where proprietary models significantly outperform open-source models?
    \item Do different modeling strategies affect a model's accuracy? Does OCR impact the reasoning accuracy of models?
    \item Does shuffling the options in questions impact model performance, indicating a potential option bias in large models?
\end{itemize}

Results for text-only questions using \textbf{\textit{Standard QA}} and \textbf{\textit{Image Only}} (implicit OCR) modeling strategy are shown in Table \ref{text-only}. For the \textbf{\textit{Standard QA}} strategy, we report results for zero shot setting in all 10 models listed  in the appendix ~\ref{ref:appx:model_hyper}. For \textbf{\textit{Image Only}} modeling, results are reported for five models that support vision or multimodal inputs for zero shot results. Columns in the table refers to different types of questions in \datasetName such as SER (series), ALP (alphabet Test), ODO (odd one out), ANA (analogy), COD (coding-decoding), NUM (number and ranking), BLR (blood relation), MTO (mathematical operations), PUZ (puzzle test), SYL (syllogisms), STC (statement and conclusions), and DAT (data sufficiency).
\begin{table*}[!htb] 
\centering
\setlength{\tabcolsep}{4pt}
\scriptsize
\scalebox{1.05}{
\begin{tabular}{l|l|cccccccccccc|c}
\toprule
&\textbf{Model} &\textbf{SER} &\textbf{ALP} &\textbf{ODO} &\textbf{ANA} &\textbf{COD} &\textbf{NUM} &\textbf{BLR} &\textbf{MTO} &\textbf{PUZ} &\textbf{SYL} &\textbf{STC} &\textbf{DAT} & Avg. Per\\
\midrule
&\multicolumn{13}{c}{\textbf{Image Only}} \\
\midrule
\multirow{5}{*}{\begin{sideways}\textbf{ {\sc Zero Shot }}\end{sideways}} 
&CogVLM-2 &14.84 &17.02 &20.00 &19.87 &24.83 &16.55 &23.81 &20.20 &20.00 &22.73 &22.12 &15.56 & 19.79\\
&InternLM-XComposer2 &18.36 &18.09 &21.76 &16.56 &17.45 &11.51 &15.87 &24.24 &25.26 &11.36 &17.31 &8.89 &17.22\\
&Qwen-VL-Chat &29.69	&23.4	&26.47&	23.84	&27.52	&23.19	&18.25	&26.26	&30.53	&6.82	&15.38	&21.59	 &22.74\\
&\textit{Gemini 1.5 Pro} &\textbf{32.42} &\textbf{31.91} &47.65 &\textbf{52.32} &27.52 &\textbf{37.41} &38.10 &29.29 &\textbf{47.37} &\textbf{47.73} &38.46 &\textbf{44.44} &\textbf{39.55}\\
&\textit{GPT-4o} &28.12 &\textbf{31.91} &49.41 &45.03 &30.87 &32.37 &\textbf{52.38} &\textbf{34.34} &36.84 &43.18 &\textbf{53.85} &33.33 &39.30\\
\midrule
\multirow{2}{*}{\rotatebox{90}{\parbox{0.5cm}{\textbf{{\sc Few \\ Shot }}}}} 
&\textit{Gemini 1.5 Pro} &23.32 &23.08 &46.11 &47.97 &24.66 &36.76 &36.59 &32.29 &42.39 &31.71 &32.67 &22.99 &33.37\\
&\textit{GPT-4o} &32.02 &29.67 &\textbf{50.30} &42.57 &\textbf{32.19} &35.29 &43.09 &25.00 &46.74 &41.46 &53.47 &34.48 &\textbf{38.85}\\
\midrule
&\multicolumn{13}{c}{\textbf{Standard QA}} \\
\midrule
\multirow{8}{*}{\begin{sideways}\textbf{ {\sc Zero Shot }}\end{sideways}} 
&Mixtral-8x7B &19.76 &19.57 &24.71 &45.52 &14.77 &26.09 &29.37 &29.59 &32.93 &24.32 &53.85 &33.33& 29.48 \\
&Llama-3 70B &35.18 &26.09 &47.65 &57.93 &36.36 &36.23 &50.79 &31.63 &60.98 &54.05 &52.88 &40.48 &44.18\\
&GPT-3.5 Turbo &35.97 &32.61 &40.00 &51.72 &36.36 &25.36 &36.51 &27.55 &46.34 &35.14 &40.38 &32.14 &36.67\\
&CogVLM-2 &22.27 &21.28 &27.65 &34.44 &22.82 &18.71 &30.95 &19.19 &29.47 &18.18 &28.85 &27.78 &25.13\\
&InternLM-XComposer2 &21.88 &24.47 &19.41 &36.42 &25.50 &28.78 &25.40 &27.27 &45.26 &40.91 &34.62 &28.89&29.90 \\
&Qwen-VL-Chat &30.08	&18.09	&23.53	&31.13	&26.85	&15.11	&24.6	&27.27	&28.26	&13.64	&15.38	&24.44 &23.19\\
&\textit{Gemini 1.5 Pro} &63.67 &39.36 &60.00 &\textbf{69.54} &\textbf{61.07} &68.35 &58.73 &45.45 &\textbf{81.05} &\textbf{65.91} &70.19 &\textbf{63.33} &\textbf{62.22}\\
&\textit{GPT-4o} &42.58 &35.11 &55.88 &65.56 &38.26 &42.45 &68.25 &41.41 &69.47 &63.64 &70.19 &43.33 &53.01\\
&LLaVA-OneVision &42.19	&32.98&	50.59&	57.62&	45.64	&36.69	&57.94&	37.37&	64.21&	50	&62.5&	46.67&	48.7\\
&Ovis1.6-Gemma2-9B &42.58	&31.91	&50	&50.99&	42.95	&46.04	&38.89	&31.31	&53.26&	27.27	&55.77	&33.33 & 42.025\\
\midrule
\multirow{6}{*}{\begin{sideways}\textbf{ {\sc Few Shot }}\end{sideways}} 
&Mixtral-8x7B &27.20 &24.72 &28.14 &50.70 &29.41 &27.41 &33.33 &29.47 &18.99 &\# &55.45 &32.10 &32.44\\
&Llama-3 70B &34.00 &16.85 &44.91 &51.41 &36.47 &34.81 &39.84 &32.63 &34.18 &\# &50.50 &34.57&37.28 \\
&GPT-3.5 Turbo &30.80 &32.58 &20.96 &47.89 &30.59 &31.11 &30.08 &29.47 &36.71 &\# &40.59 &34.57&33.21 \\
&\textit{Gemini 1.5 Pro} &\textbf{63.24} &37.36 &\textbf{59.28} &68.92 &60.27 &\textbf{68.38} &58.54 &43.75 &80.43 &63.41 &70.30 &62.07 &\textbf{61.32}\\
&\textit{GPT-4o} &42.29 &\textbf{40.66} &58.08 &67.57 &44.52 &40.44 &\textbf{69.92} &\textbf{46.88} &72.83 &63.41 &\textbf{71.29} & * &56.17\\
\midrule
&\multicolumn{13}{c}{\textbf{Advanced Reasoning Models}} \\
\midrule
\multirow{1}{*}{\begin{sideways}\textbf{ }\end{sideways}} 
&OpenAI o1-preview &	80.62 &	90.22	&84.05	&73.13	& 85 &	85.83 &	83.61 &	83.7 &	84.81 &	81.08	&72.28 &	83.33	& \bf 81.88\\
\bottomrule
\end{tabular}
}
\caption{\textbf{Text Only Question Results.} Zero-shot and few-shot performance of different models across various \textit{text-only} categories. We report results using two  modeling strategies \textit{Image Only} and \textit{Standard QA}. \textit{italics} font for propriety models, i.e., money or API access is required to run these models. The \# is due to the category's solution contains images thus restricting few shot on text-only models.  Note: (*) In some models, a common issue arises when a model refrains
from providing a response due to safety concerns, often stemming from misinterpretation of the image’s intent, e.g., thinking it as CAPTCHA.}
\label{text-only}
\end{table*}
\begin{table*}[!htb] 
\centering
\scriptsize
\setlength{\tabcolsep}{3pt} 
\scalebox{1.03}{
\begin{tabular}{l|l|rrrrrrrrrrrrrr|c}
\toprule
&\textbf{Model} &\textbf{DIR} &\textbf{VEN} &\textbf{TIM} &\textbf{MIS} &\textbf{NVS} &\textbf{NVO} &\textbf{NVA} &\textbf{INC} &\textbf{MIR} &\textbf{CUB} &\textbf{PAP} &\textbf{EMB} &\textbf{FIG} &\textbf{DOT} & Avg. Prec. \\
\midrule
&\multicolumn{15}{c}{\textbf{Interleaved}} \\
\midrule
\multirow{4}{*}{\rotatebox{90}{\parbox{0.5cm}{\textbf{{\sc Zero \\ Shot }}}}}
&Qwen-VL-Chat	&28.12	&19.82	&19.61	&12.6	&22.11	&27.14	&22.00	&23.40	&27.17	&15.73	&23.96	&30.21	&8.45	&17.39 & 21.26\\
&\textit{Gemini 1.5 Pro} &\textbf{63.54} &\textbf{64.86} &\textbf{70.59} &\textbf{37.01} &\textbf{33.68} &25.71 &\textbf{32} &\textbf{38.3} &\textbf{35.87} &\textbf{43.82} &30.21 &36.46 &\textbf{46.48} &\textbf{30.43} & \textbf{42.06} \\
&\textit{GPT-4o} &37.50 &50.45 &41.18 &29.92 &16.84 &22.86 &26 &23.4 &34.78 &35.96 &27.08 &22.92 &45.07 &17.39 &30.81 \\
&LLaVA-OneVision &27.27	&39.64	&44.44	&32	&14.74	&\textbf{28.57}	&26	&26.6	&32.61	&36.59	&26.04	&\textbf{37.5}	& 33.8 &	26.09 &	30.85\\
&Ovis1.6-Gemma2-9B &35.42	&36.04&	39.22&	23.62&	25.26&	28.57&	19&	32.98&	32.61&	10.11&	29.17&	23.96&	9.86&	21.74&	26.25\\
\midrule
\multirow{2}{*}{\rotatebox{90}{\parbox{0.5cm}{\textbf{{\sc Few \\ Shot }}}}}
&\textit{Gemini 1.5 Pro} &62.37 &63.89 &68.75 &36.29 &31.52 &23.88 &29.9 &36.26 &33.71 &41.86 &27.96 &34.41 &44.12 &20 & \textbf{39.63}\\
&\textit{GPT-4o} &39.78 &52.78 &52.08 &27.42 &17.39 &* &* &19.78 &* &38.37 &\textbf{33.33} &* &41.18 &* & 35.79\\
\midrule
&\multicolumn{15}{c}{\textbf{Image Only}} \\
\midrule
\multirow{5}{*}{\begin{sideways}\textbf{ {\sc Zero Shot }}\end{sideways}} 
&CogVLM-2 &18.75 &18.02 &25.49 &14.96 &18.95 &20 &8.00 &12.77 &7.61 &19.10 &16.67 &12.50 &12.68 &4.35 &14.98\\
&Qwen-VL-Chat &21.05 &26.13 &27.45 &22.22 &\textbf{26.32} &21.43 &17.00 &21.28 &19.57 &25.84 &25 &18.75 &18.31 &17.39 &21.98\\
&InternLM-XComposer2 &20.83 &20.72 &15.69 &17.32 &15.79 &11.43 &10.00 &14.89 &8.70 &19.10 &10.42 &11.46 &22.54 &8.70 &14.82\\
&\textit{Gemini 1.5 Pro} &\textbf{52.08} &\textbf{37.84} &\textbf{49.02} &25.2 &24.21 &24.29 &27  &\textbf{26.6} &\textbf{29.35} &32.58 &23.96 &23.96 &\textbf{42.25} &\textbf{34.78} & \textbf{32.36}\\
&\textit{GPT-4o} &40.62 &31.53 &33.33 &22.05 &22.11 &\textbf{25.71} &19 &24.47 &23.91 &26.97 &\textbf{34.38} &23.96 &\textbf{42.25} &21.74 & 28.00\\
\midrule
\multirow{2}{*}{\rotatebox{90}{\parbox{0.5cm}{\textbf{{\sc Few \\ Shot }}}}} 
&\textit{Gemini 1.5 Pro} &47.31 &27.78 &33.33 &\textbf{29.03} &25.00 &23.88 &21.65 &23.08 &21.35 &\textbf{37.21} &32.26 &19.35 &22.06 &25.00 &\textbf{27.73} \\
&\textit{GPT-4o} &31.18 &29.63 &37.50 &22.58 &23.91 &14.93 &\textbf{23.71} &21.98 &21.35 &23.26 &26.88 &\textbf{26.88} &39.71 &20.00 & 25.96\\
\midrule
&\multicolumn{15}{c}{\textbf{Standard VQA}} \\
\midrule
\multirow{4}{*}{\begin{sideways}\textbf{ {\sc Zero Shot }}\end{sideways}} 
&CogVLM-2 &15.62 &12.61 &29.41 &11.02 &8.42 &4.29 &6 &3.19 &11.96 &15.73 &9.38 &10.42 &8.45 &17.39 & 11.70\\
&Qwen-VL-Chat &21.88 &18.92 &27.45 &5.51 &23.16 &22.86 &20 &24.47 &\textbf{26.09} &8.99 &20.83 &19.79 &8.45 &8.7 & 18.36 \\
&InternLM-XComposer2 &25 &20.72 &25.49 &17.32 &18.95 &8.57 &15 &5.32 &16.3 &12.36 &20.83 &10.42 &12.68 &13.04 & 15.85 \\
&\textit{Gemini 1.5 Pro} &54.17 &\textbf{49.55} &62.75 &\textbf{37.8} &24.21 &24.29 &21 &\textbf{29.79} &21.74 &\textbf{46.07} &23.96 &23.96 &40.85 &\textbf{26.09} & \textbf{34.73} \\
&\textit{GPT-4o} &50 &45.95 &39.22 &28.35 &\textbf{32.63} &\textbf{25.71} &\textbf{26} &18.09 &22.83 &40.45 &23.96 &\textbf{28.12} &40.85 &\textbf{26.09} & 32.01 \\
\midrule
\multirow{2}{*}{\rotatebox{90}{\parbox{0.5cm}{\textbf{{\sc Few \\ Shot }}}}} 
&\textit{Gemini 1.5 Pro} &\textbf{61.29} &47.22 &\textbf{68.75} &32.26 &17.39 &16.42 &18.56 &27.47 &20.22 &44.19 &20.43 &25.81 &\textbf{44.12} &25 & \textbf{33.50} \\
&\textit{GPT-4o} &41.94 &49.07 &45.83 &27.42 &15.22 &23.88 &22.68 &15.38 &25.84 &34.88 &\textbf{26.88} &22.58 &35.29 &25 &29.42 \\
\bottomrule
\end{tabular}
}
\caption{\textbf{Multi-modality Question Results}: Zero-shot and few-shot performance of different VLMs across various Text+Vision categories. We report results using all three different modelling strategies proposed to handle multimodality data, i.e., \textit{Interleaved}, \textit{Image Only} and \textit{Standard VQA}. \textit{italics} font for proprietary models, i.e., money or API access is required to run these models. Note: (*) In some models, a common issue arises when a model refrains from providing a response due to safety concerns, often stemming from misinterpretation of the image's intent, e.g., thinking it as CAPTCHA.}
\label{text-and-vision}
\end{table*}

Results for the multimodality questions are reported in Table \ref{text-and-vision}. Results are reported using three modeling strategies listed in the section above namely \textbf{\textit{Interleaved, Image Only }} and \textbf{\textit{ Standard VQA}}.  Columns in the table refer to different categories of visual text reasoning questions in \datasetName such as DIR (direction sense), VEN (venn diagrams), TIM (time and clock), MIS (missing character), NVS (non-verbal series), NVO (non-verbal odd one out), NVA (non-verbal analogy), INC (incomplete figure), MIR (mirror, water and images), CUB (cube and dice), PAP (paper folding and cutting), EMB (embedded figure), FIG (figure partition), DOT (dot problem). Option bias ablation results for Gemini-1.5-pro and GPT-4o models are shown in Appendix Tables \ref{tab:option_shuff_gemini_text}, ~\ref{tab:option_shuff_gemini_vis}, ~\ref{tab:option_shuff_gpt_text}, and ~\ref{tab:option_shuff_gpt_vis}.
\textbf{Proprietary models outperform open-source models.} From results in Table \ref{text-only} and \ref{text-and-vision}, we can observe that proprietary models, such as Gemini Pro 1.5,  GPT-4o and o1-preview, outperform other open-source models in nearly every question category, especially o1-preview, which operates by utilizing internal chain-of-thought reasoning, allowing it to analyze complex problems step-by-step before arriving at a solution. Proprietary models demonstrate nearly double the accuracy of open-source models on \datasetName questions, both in text-based and multimodal tasks, across all modeling strategies outlined in the previous section. Notably, Gemini Pro 1.5 consistently outperforms GPT-4o across most strategies, excelling in both text and multimodal question types.

For open-source models, LLaVA-OneVision-72b-ov-chat excels in text and multimodality tasks, achieving SOTA results on tasks such as Non-verbal Odd One Out and Embedded Figure, also matching or surpassing proprietary models in several other categories. Ovis1.6-Gemma2-9B, although smaller, outperforms GPT-4o and LLaVA-OneVision in several vision categories, but its overall performance is still lower.

\textbf{Modeling Strategy is important.} For text-only questions, the \textbf{\textit{Standard QA}} strategy clearly outperforms \textbf{\textit{Image Only}} modeling. Introducing the burden of doing OCR on top of reasoning tends to confuse models or exacerbate the difficulty of the task. This effect is particularly noticeable with smaller open-source models, which struggle to accurately extract characters and integrate them into context. However, proprietary models such as GPT-4o and Gemini-Pro still achieve superior results using \textbf{\textit{Image Only}} compared to open-source models employing \textbf{\textit{Standard QA}} or text-only processing alone. Few-shot results present a mixed picture: where as models like Mixtral and GPT-4o show improved performance with added exemplars, others experience a significant decline. Tables \ref{text-only} and \ref{text-and-vision} show that proprietary models generally experience a smaller performance drop than open-source models in such scenarios.

\textbf{\textit{Interleaving}} text and images performs better than \textbf{\textit{Standard VQA}} and  \textbf{\textit{Image Only}} strategy for most categories. This underscores the importance of presenting text and images separately and in a more detailed manner, providing appropriate context, and treating the image and text as distinct entities. Our results in Table \ref{text-and-vision} show that this approach significantly improves VLM results, as demonstrated by the superior performance of open-source models such as LLaVA-OneVision-72b and Ovis-9b, which handle interleaved text and images more effectively than others.
Table \ref{text-and-vision} also shows that few-shot prompting consistently underperforms compared to zero-shot for all VLM models, suggesting that adding exemplars may confuse VLMs, hindering their focus on logical reasoning.

\textbf{Multimodal reasoning is significantly harder.} Comparing the best and worst performing models in Tables \ref{text-only} and \ref{text-and-vision}, it is evident that multimodal reasoning is considerably more challenging than textual reasoning for current state-of-the-art models.
Multimodal questions see less than 45\% accuracy, whereas the best model for textual reasoning, o1-preview, exceeds 80\%. Even fast models such as Gemini 1.5 pro achieve over 60\% accuracy in textual reasoning, highlighting the gap between VLMs and LLMs and the need for better architectures and datasets for VLMs.

\textbf{Question category analysis}. The results from Table \ref{text-only} reveal that even though LLMs generally perform better on the text-only subset of \datasetName, the high standard deviation (11.04) indicates significant variability in model performance across different question types. This variability may stem from some overlap between \datasetName and other open-source datasets, suggesting that there are still areas where LLMs exhibit limitations in reasoning capabilities. This is especially evident in Alphabet Test (ALP) category and Mathematical Operations (MTO) category, where the accuracy is more than two standard deviations away from the mean accuracy of the model. This could also be attributed to the potential difficulty of these question types; however, that analysis has been left for future work.

VLMs have shown poor performance across all categories for multimodal questions, with the best-performing model achieving correct answers only 42\% of the time. Even the standard deviation for accuracy of the best model is 9, indicating that VLMs struggle more with certain question categories than others. Specifically, we observe that VLMs perform notably poorly on categories such as DOT (Dot Problems), NVS (Non-verbal Series), and NVO (Non-verbal Odd One Out). These categories require identifying correlations or patterns between multiple images or recognizing emerging patterns in a sequence of images. This task is akin to identifying similar and evolving patterns in different parts of an image and predicting the next possible pattern. Although vision models excel at recognizing existing patterns, they struggle with predicting new patterns. 

\textbf{\datasetName presents a challenging task for SOTA LLMs and VLMs}. Based on the findings in Tables \ref{text-only} and \ref{text-and-vision}, it is evident that the proposed dataset presents a challenging task for all state-of-the-art LLM and VLM models. None of the open-source models achieve accuracy exceeding 50\% on text-only questions and 35\% on multimodal questions, with proprietary models achieving 62\% and 42\% accuracy, respectively. Although o1-preview is categorized in a separately, it got more than 80\% accuracy on text-only data, it cannot do multimodal reasoning. Many of the models tested did not even reach random selection/guess baseline of 24.52\% (261 question had 5 options and 2467 question has 4 options), indicating that current LLMs and VLMs struggle with cognitive reasoning questions.

\textbf{Human vs Model.} Preliminary human evaluation results shown in appendix section \ref{sec:appx:human_eval} show that average human accuracy for textual and multi-modal questions is more than 80\%, much greater than 62\% (text) and 42\% (visual) for best performing propriety model. Although the o1-preview model have achieved near human accuracy on textual reasoning, it is significantly slower than other models. These findings suggest potential for future advances in tackling diverse textual and visual reasoning questions.

\textbf{Options Bias Ablation.} From Gemini 1.5 Pro Results in Table ~\ref{tab:option_shuff_gemini_text} and ~\ref{tab:option_shuff_gemini_vis} we can see that aggregate performance for both multimodal and text question is affected by the position of the correct option (variation in performance ranging from -4 to +6 percent when compared to random for text and -5 to +5 percent for multimodal categories). We can make a similar observation on GPT-4o (Tables ~\ref{tab:option_shuff_gpt_text} and ~\ref{tab:option_shuff_gpt_vis}); however, GPT-4o has smaller variation in performance than Gemini Pro. These variations suggest that the model's responses to logical questions may be either memorized or guessed, indicating a potential limitation in LLMs and VLMs. Even o1-preview, as shown in Table \ref{tab:option_shuff_o1_text}, exhibits performance variations akin to those of GPT-4o, suggesting that advanced reasoning models are similarly affected by bias. 

\begin{figure}[!htb]
    \centering
    \includegraphics[width=1\linewidth]{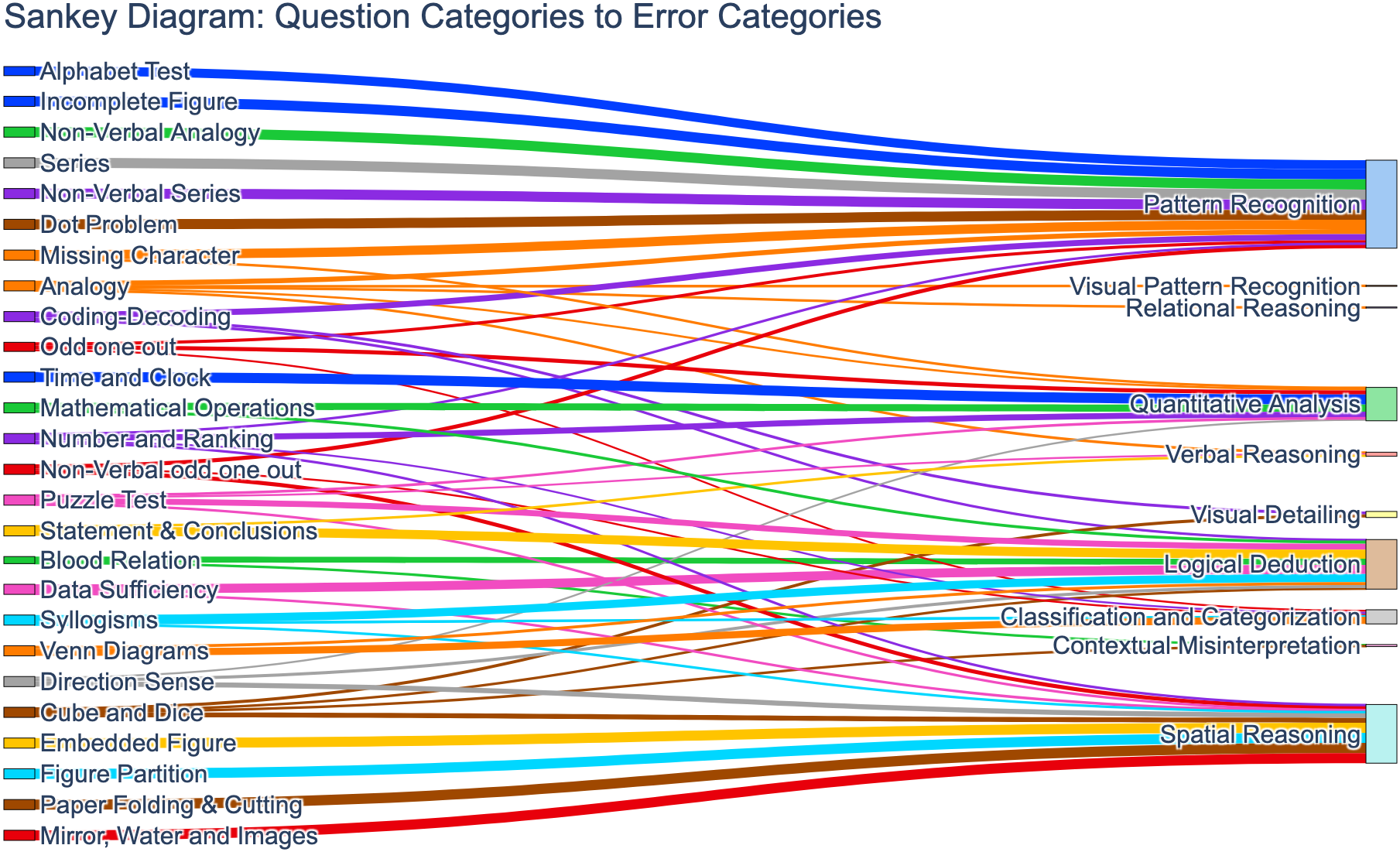}
    \caption{\textbf{Error Analysis.} Overview of errors Gemini 1.5 Pro makes across different question categories.}
    \label{fig:error_analysis}
    \vspace{-0.25em}
\end{figure}

\textbf{Error Analysis.}
We manually conducted an error analysis of 260 questions (10 from each question category) for Gemini 1.5 pro, and we identified distinct patterns in reasoning and error categorization. We have categorised errors based on the cognitive dimensions outlined in section~\ref{sec:ntse_benchmark}. The Sankey diagram in Figure~\ref{fig:error_analysis} illustrates how errors across various question categories correspond to specific error types. 
A key observation is that many errors arise from \textbf{Pattern Recognition} failures, especially in categories like \emph{Alphabet Tests, Non-verbal Analogy, and Series} questions, where the model struggled with recurring patterns and sequence shifts, highlighting challenges in complex pattern-based reasoning.
We also noted frequent errors in \textbf{Spatial Reasoning} and \textit{Logical Deduction} tasks, particularly in spatial or diagrammatic questions such as \textit{Cube and Dice, Embedded Figure, and Paper Folding and Cutting}. These questions often require pattern recognition, shape manipulation, or deducing logical relations from limited visual data. The figure also shows that errors in \textbf{Quantitative Analysis} were common in numerical questions such as \textit{Time and Clock} and \textit{Mathematical Operations}, indicating the model excels in simpler tasks but struggles with complex number sequences and operations. The error distribution reveals key insights into the model's strengths and weaknesses, guiding future improvements. A detailed error analysis for each category is included in Appendix ~\ref{sec:appx:reason_analysis}.

\section{Related works}
\textbf{Texual and Multimodal Reasoning datasets.} There exist multiple datasets to test domain specific textual (math, science, medicine) QA and reasoning abilities of LLMs and VLMs knowledge such as SciBench \cite{wang2023scibench}, SciEval\cite{sun2024scieval}, MMMU \cite{yue2024mmmu}, MathVista \cite{lu2023mathvista}, JEEbench \cite{arora2023have}, MathVerse \cite{zhang2024mathverse}, OlympiadBench \cite{he2024olympiadbench} and many others based on real-world images or other domains \cite{he2020pathvqa,soni2022radqa,liu2023mmbench,thrush2022winoground,li2023seed}. Even most existing visual and multimodal reasoning datasets in the literature are domain-specific M3Exam \cite{zhang2024m3exam}, RAVEN \cite{zhang2019raven}, or they involve reasoning about real-world images \cite{liu2023mmbench,wang2024exploring,thrush2022winoground,li2023seed} with basic common sense reasoning questions such as CLEVR \cite{johnson2017clevr}. However, current research has not thoroughly explored the capabilities of large models in addressing cognitive/critical reasoning problems for both textual and multimodal data. \datasetName is different from all already existing datasets in the literature because it explicitly focuses on testing cognitive reasoning abilities of large deep learning models.
Although benchmarks like BBH \cite{suzgun2022challenging} focus on text-based logical deduction and direction sense, \datasetName offers a more holistic assessment by integrating both textual and visual elements. For example, it includes categories that combine text with visual answers, such as Venn diagrams and figure partition questions, which BBH does not cover.
\datasetName also includes areas overlooked by other benchmarks, such as series, coding-decoding, and blood relations. With 2,728 multiple-choice questions, it provides a broader evaluation than the 770 puzzles in BBH.

Furthermore, benchmarks such as \cite{jiang2024marvel} emphasize visual abstraction and reasoning but lack the integration of text and visual elements found in \datasetName. Our dataset also goes beyond Raven’s Progressive Matrices (RPMs) \cite{zhang2019raven} and related visual analogy problems, which are central to \citet{malkinski2023review}, by incorporating a wider array of cognitive tasks. This includes diverse categories such as coding-decoding, number and ranking, blood relations, and mathematical operations. \datasetName provides a fine-grained analysis of model performance across 26 distinct problem categories and investigates various modeling strategies, setting it apart from other benchmarks. 
Datasets such as \cite{wang2024exploring,liu2023mmbench,masry-etal-2022-chartqa} explore mathematical reasoning in visual contexts but do not permit multiple images in the same prompt or reasoning across images. This is crucial for identifying patterns in categories of \datasetName, such as Paper Folding and Cutting, Embedded Figure, Figure Partition, and Mirror/Water images. 

Although ARC \cite{chollet2019measure} also targets general human cognitive understanding, its questions are primarily focused on visual tasks such as pattern completion, interpolation, denoising, and simple object counting. These tasks are relatively straightforward and often lack the need for complex reasoning. 
            By leveraging NTSE exam questions, our dataset offers real-world relevance and facilitates quantitative evaluation through a multiple-choice format. Additionally, NTSEBench enables fine-grained performance analysis across cognitive dimensions, emphasizing knowledge integration and reasoning, which more closely mirrors human cognitive processes. These attributes make NTSEBench a more comprehensive and effective tool for evaluating AI's cognitive capabilities.

Language Writ Large \cite{harnad2024language} delves into the theoretical foundations of large language models (LLMs) like ChatGPT, offering insights into their unexpected capabilities through the analysis of technology dialogues. In contrast, NTSEBench provides quantitative performance metrics for LLMs across a wide range of reasoning tasks. 
MMMU \cite{yue2024mmmu}, on the other hand, spans six disciplines and 30 college subjects, providing a broad overview of subject-specific knowledge. In contrast, NTSEBench offers a deeper, more focused evaluation of cognitive reasoning skills, making it an essential tool for assessing the reasoning capabilities of LLMs.

\textbf{Zero shot and few shot prompt engineering for textual and multimodal input.} Our work is also related to prompting design and prompt engineering for LLMs \cite{brown2020language,chen2023unleashing,gupta2023multi,khot2022decomposed,wei2022chain,sahoo2024systematic,ali2024prompt} and VLMs \cite{xu2024collage,bai2023qwen,liu2024visual,dai2024instructblip}. There are numerous studies on multimodal and vision Chain-of-Thought (CoT) prompting \cite{zhang2023multimodal, shao2024visual}. These studies overlook the ability of state-of-the-art vision-language models (VLMs) to process multiple images simultaneously, a limitation of older models with smaller context windows. Newer models can handle few-shot examples for CoT prompting, and we explore three prompting strategies enabled by their extended context windows.

\section{Conclusion and Future Work}

We developed a new dataset, \datasetName, to assess the advanced analytical and logical reasoning capabilities of large deep learning models (LLMs and VLMs).  We also propose four distinct modeling strategies for handling multimodal data (text and images) across different question types in \datasetName. These strategies enable us to conduct a fair and comprehensive comparison between proprietary and open-source models using both zero-shot and few-shot scenarios. 
Our findings show that both LLMs and VLMs struggle with advanced visual reasoning tasks, with VLMs performing worse on multi-modal questions than LLMs on textual ones. Proprietary models also consistently outperform open-source models, correctly predicting twice as many questions.
Overall, our results underscore that \datasetName poses significantly greater challenges for state-of-the-art LLMs and VLMs. \textbf{Future directions.} (a) Our results indicated that VLM models have difficulty predicting novel patterns, implying that addressing this challenge may involve either architectural modifications or the integration of generative models alongside VLMs for these question types. (b) Given the limited data available for cognitive reasoning questions, we plan to use data augmentation strategies to increase the number of samples.
\newpage
\section*{Limitations}
While our dataset is new and sourced for different sources when compared to datasets already present in the literature, there still might be some overlap in reasoning questions, especially in textual reasoning. So all the dataset instances might not be independent and exclusive. This dataset is solely created in English, so no other languages are represented; therefore, we cannot analyze whether language variations can have a significant impact on the reasoning capabilities of these large models. Our modelling strategies were limited to zero-shot and few-shot COT prompting. We did not evaluate whether fine-tuning these large models on a few examples from each of the categories would further improve results. This was due to the limitation of both GPU resources and large cost of fine-tuning for proprietary models. Finally, our human evaluation study involved only three annotators with undergraduate degrees, which may limit the generalizability of the results to the broader human population. We plan to address all these limitations in the future extension of this work.

\section*{Ethics Statement}
As authors of this work, we affirm that our research adheres to the highest ethical standards in both its conduct and publication. We have carefully considered and addressed various ethical considerations inherent in computational linguistics methodologies to ensure responsible and fair practices. We prioritize transparency by providing detailed information to facilitate the reproducibility of our results. This includes sharing our code and datasets, which are sourced from publicly available datasets and handled in compliance with ethical guidelines established by the original authors. Although our paper accurately reflects our experimental findings, we acknowledge the inherent variability associated with large language models, which we mitigate by maintaining a fixed temperature setting. We provide comprehensive descriptions of our annotations, dataset splits, models used, and prompting methods employed, ensuring the reproducibility of our work. For grammar correction, we utilize AI-based writing assistants, and for coding, we leverage tools such as Copilot. Importantly, the genesis of our research ideas and the execution of our study were entirely independent of AI assistance.

\section*{Acknowledgements}
Research was sponsored by the Army Research Office and was accomplished under Grant Number W911NF-20-1-0080. The views and conclusions contained in this document are those of the authors and should not be interpreted as representing the official policies, either expressed or implied, of the Army Research Office or the U.S. Government. The U.S. Government is authorized to reproduce and distribute reprints for Government purposes notwithstanding any copyright notation herein. This work was partially funded by ONR Contract N00014-23-1-2364. We extend our gratitude to the annotators who verified our data extraction and corresponding question answer pairs. We extend our sincere appreciation to Jennifer Sheffield from the University of Pennsylvania for her administrative support. Lastly, we extend our appreciation to the reviewing team for their insightful comments.
\bibliography{anthology,custom}

\begin{thebibliography}{56}
\expandafter\ifx\csname natexlab\endcsname\relax\def\natexlab#1{#1}\fi

\bibitem[{Achiam et~al.(2023)Achiam, Adler, Agarwal, Ahmad, Akkaya, Aleman, Almeida, Altenschmidt, Altman, Anadkat et~al.}]{achiam2023gpt}
Josh Achiam, Steven Adler, Sandhini Agarwal, Lama Ahmad, Ilge Akkaya, Florencia~Leoni Aleman, Diogo Almeida, Janko Altenschmidt, Sam Altman, Shyamal Anadkat, et~al. 2023.
\newblock Gpt-4 technical report.
\newblock \emph{arXiv preprint arXiv:2303.08774}.

\bibitem[{AI@Meta(2024)}]{llama3modelcard}
AI@Meta. 2024.
\newblock \href {https://github.com/meta-llama/llama3/blob/main/MODEL_CARD.md} {Llama 3 model card}.

\bibitem[{Ali et~al.(2024)Ali, Li, Yang, Cheng, Cao, Huang, Hu, Yu, and Wang}]{ali2024prompt}
Muhammad~Asif Ali, Zhengping Li, Shu Yang, Keyuan Cheng, Yang Cao, Tianhao Huang, Lijie Hu, Lu~Yu, and Di~Wang. 2024.
\newblock Prompt-saw: Leveraging relation-aware graphs for textual prompt compression.
\newblock \emph{arXiv preprint arXiv:2404.00489}.

\bibitem[{Arora et~al.(2023)Arora, Singh et~al.}]{arora2023have}
Daman Arora, Himanshu~Gaurav Singh, et~al. 2023.
\newblock Have llms advanced enough? a challenging problem solving benchmark for large language models.
\newblock \emph{arXiv preprint arXiv:2305.15074}.

\bibitem[{Bai et~al.(2023)Bai, Bai, Yang, Wang, Tan, Wang, Lin, Zhou, and Zhou}]{bai2023qwen}
Jinze Bai, Shuai Bai, Shusheng Yang, Shijie Wang, Sinan Tan, Peng Wang, Junyang Lin, Chang Zhou, and Jingren Zhou. 2023.
\newblock Qwen-vl: A frontier large vision-language model with versatile abilities.
\newblock \emph{arXiv preprint arXiv:2308.12966}.

\bibitem[{Brown et~al.(2020)Brown, Mann, Ryder, Subbiah, Kaplan, Dhariwal, Neelakantan, Shyam, Sastry, Askell, Agarwal, Herbert-Voss, Krueger, Henighan, Child, Ramesh, Ziegler, Wu, Winter, Hesse, Chen, Sigler, Litwin, Gray, Chess, Clark, Berner, McCandlish, Radford, Sutskever, and Amodei}]{brown2020language}
Tom~B. Brown, Benjamin Mann, Nick Ryder, Melanie Subbiah, Jared Kaplan, Prafulla Dhariwal, Arvind Neelakantan, Pranav Shyam, Girish Sastry, Amanda Askell, Sandhini Agarwal, Ariel Herbert-Voss, Gretchen Krueger, Tom Henighan, Rewon Child, Aditya Ramesh, Daniel~M. Ziegler, Jeffrey Wu, Clemens Winter, Christopher Hesse, Mark Chen, Eric Sigler, Mateusz Litwin, Scott Gray, Benjamin Chess, Jack Clark, Christopher Berner, Sam McCandlish, Alec Radford, Ilya Sutskever, and Dario Amodei. 2020.
\newblock \href {http://arxiv.org/abs/2005.14165} {Language models are few-shot learners}.

\bibitem[{Bubeck et~al.(2023)Bubeck, Chandrasekaran, Eldan, Gehrke, Horvitz, Kamar, Lee, Lee, Li, Lundberg et~al.}]{bubeck2023sparks}
S{\'e}bastien Bubeck, Varun Chandrasekaran, Ronen Eldan, Johannes Gehrke, Eric Horvitz, Ece Kamar, Peter Lee, Yin~Tat Lee, Yuanzhi Li, Scott Lundberg, et~al. 2023.
\newblock Sparks of artificial general intelligence: Early experiments with gpt-4.
\newblock \emph{arXiv preprint arXiv:2303.12712}.

\bibitem[{Chen et~al.(2023)Chen, Zhang, Langren{\'e}, and Zhu}]{chen2023unleashing}
Banghao Chen, Zhaofeng Zhang, Nicolas Langren{\'e}, and Shengxin Zhu. 2023.
\newblock Unleashing the potential of prompt engineering in large language models: a comprehensive review.
\newblock \emph{arXiv preprint arXiv:2310.14735}.

\bibitem[{Chollet(2019)}]{chollet2019measure}
Fran{\c{c}}ois Chollet. 2019.
\newblock On the measure of intelligence.
\newblock \emph{arXiv preprint arXiv:1911.01547}.

\bibitem[{Dai et~al.(2024)Dai, Li, Li, Tiong, Zhao, Wang, Li, Fung, and Hoi}]{dai2024instructblip}
Wenliang Dai, Junnan Li, Dongxu Li, Anthony Meng~Huat Tiong, Junqi Zhao, Weisheng Wang, Boyang Li, Pascale~N Fung, and Steven Hoi. 2024.
\newblock Instructblip: Towards general-purpose vision-language models with instruction tuning.
\newblock \emph{Advances in Neural Information Processing Systems}, 36.

\bibitem[{Dong et~al.(2024)Dong, Zhang, Zang, Cao, Wang, Ouyang, Wei, Zhang, Duan, Cao et~al.}]{dong2024internlm}
Xiaoyi Dong, Pan Zhang, Yuhang Zang, Yuhang Cao, Bin Wang, Linke Ouyang, Xilin Wei, Songyang Zhang, Haodong Duan, Maosong Cao, et~al. 2024.
\newblock Internlm-xcomposer2: Mastering free-form text-image composition and comprehension in vision-language large model.
\newblock \emph{arXiv preprint arXiv:2401.16420}.

\bibitem[{Dubey et~al.(2024)Dubey, Jauhri, Pandey, Kadian, Al-Dahle, Letman, Mathur, Schelten, Yang, Fan et~al.}]{dubey2024llama}
Abhimanyu Dubey, Abhinav Jauhri, Abhinav Pandey, Abhishek Kadian, Ahmad Al-Dahle, Aiesha Letman, Akhil Mathur, Alan Schelten, Amy Yang, Angela Fan, et~al. 2024.
\newblock The llama 3 herd of models.
\newblock \emph{arXiv preprint arXiv:2407.21783}.

\bibitem[{Fujitake(2024)}]{fujitake2024dtrocr}
Masato Fujitake. 2024.
\newblock Dtrocr: Decoder-only transformer for optical character recognition.
\newblock In \emph{Proceedings of the IEEE/CVF Winter Conference on Applications of Computer Vision}, pages 8025--8035.

\bibitem[{Gupta et~al.(2023)Gupta, Pandya, Kataria, Gupta, and Roth}]{gupta2023multi}
Vatsal Gupta, Pranshu Pandya, Tushar Kataria, Vivek Gupta, and Dan Roth. 2023.
\newblock Multi-set inoculation: Assessing model robustness across multiple challenge sets.
\newblock \emph{arXiv preprint arXiv:2311.08662}.

\bibitem[{Harnad(2024)}]{harnad2024language}
Stevan Harnad. 2024.
\newblock Language writ large: Llms, chatgpt, grounding, meaning and understanding.
\newblock \emph{arXiv preprint arXiv:2402.02243}.

\bibitem[{He et~al.(2024)He, Luo, Bai, Hu, Thai, Shen, Hu, Han, Huang, Zhang et~al.}]{he2024olympiadbench}
Chaoqun He, Renjie Luo, Yuzhuo Bai, Shengding Hu, Zhen~Leng Thai, Junhao Shen, Jinyi Hu, Xu~Han, Yujie Huang, Yuxiang Zhang, et~al. 2024.
\newblock Olympiadbench: A challenging benchmark for promoting agi with olympiad-level bilingual multimodal scientific problems.
\newblock \emph{arXiv preprint arXiv:2402.14008}.

\bibitem[{He et~al.(2020)He, Zhang, Mou, Xing, and Xie}]{he2020pathvqa}
Xuehai He, Yichen Zhang, Luntian Mou, Eric Xing, and Pengtao Xie. 2020.
\newblock Pathvqa: 30000+ questions for medical visual question answering.
\newblock \emph{arXiv preprint arXiv:2003.10286}.

\bibitem[{Jiang et~al.(2024{\natexlab{a}})Jiang, Sablayrolles, Roux, Mensch, Savary, Bamford, Chaplot, Casas, Hanna, Bressand et~al.}]{jiang2024mixtral}
Albert~Q Jiang, Alexandre Sablayrolles, Antoine Roux, Arthur Mensch, Blanche Savary, Chris Bamford, Devendra~Singh Chaplot, Diego de~las Casas, Emma~Bou Hanna, Florian Bressand, et~al. 2024{\natexlab{a}}.
\newblock Mixtral of experts.
\newblock \emph{arXiv preprint arXiv:2401.04088}.

\bibitem[{Jiang et~al.(2024{\natexlab{b}})Jiang, Zhang, Sun, Sourati, Ahrabian, Ma, Ilievski, and Pujara}]{jiang2024marvel}
Yifan Jiang, Jiarui Zhang, Kexuan Sun, Zhivar Sourati, Kian Ahrabian, Kaixin Ma, Filip Ilievski, and Jay Pujara. 2024{\natexlab{b}}.
\newblock Marvel: Multidimensional abstraction and reasoning through visual evaluation and learning.
\newblock \emph{arXiv preprint arXiv:2404.13591}.

\bibitem[{Johnson et~al.(2017)Johnson, Hariharan, Van Der~Maaten, Fei-Fei, Lawrence~Zitnick, and Girshick}]{johnson2017clevr}
Justin Johnson, Bharath Hariharan, Laurens Van Der~Maaten, Li~Fei-Fei, C~Lawrence~Zitnick, and Ross Girshick. 2017.
\newblock Clevr: A diagnostic dataset for compositional language and elementary visual reasoning.
\newblock In \emph{Proceedings of the IEEE conference on computer vision and pattern recognition}, pages 2901--2910.

\bibitem[{Khot et~al.(2022)Khot, Trivedi, Finlayson, Fu, Richardson, Clark, and Sabharwal}]{khot2022decomposed}
Tushar Khot, Harsh Trivedi, Matthew Finlayson, Yao Fu, Kyle Richardson, Peter Clark, and Ashish Sabharwal. 2022.
\newblock Decomposed prompting: A modular approach for solving complex tasks.
\newblock \emph{arXiv preprint arXiv:2210.02406}.

\bibitem[{King(2023)}]{king2023administration}
Michael King. 2023.
\newblock Administration of the text-based portions of a general iq test to five different large language models.
\newblock \emph{Authorea Preprints}.

\bibitem[{Li et~al.(2024)Li, Zhang, Guo, Zhang, Li, Zhang, Zhang, Li, Liu, and Li}]{li2024llava}
Bo~Li, Yuanhan Zhang, Dong Guo, Renrui Zhang, Feng Li, Hao Zhang, Kaichen Zhang, Yanwei Li, Ziwei Liu, and Chunyuan Li. 2024.
\newblock Llava-onevision: Easy visual task transfer.
\newblock \emph{arXiv preprint arXiv:2408.03326}.

\bibitem[{Li et~al.(2023)Li, Wang, Wang, Ge, Ge, and Shan}]{li2023seed}
Bohao Li, Rui Wang, Guangzhi Wang, Yuying Ge, Yixiao Ge, and Ying Shan. 2023.
\newblock Seed-bench: Benchmarking multimodal llms with generative comprehension.
\newblock \emph{arXiv preprint arXiv:2307.16125}.

\bibitem[{Liu et~al.(2024)Liu, Li, Wu, and Lee}]{liu2024visual}
Haotian Liu, Chunyuan Li, Qingyang Wu, and Yong~Jae Lee. 2024.
\newblock Visual instruction tuning.
\newblock \emph{Advances in neural information processing systems}, 36.

\bibitem[{Liu et~al.(2023)Liu, Duan, Zhang, Li, Zhang, Zhao, Yuan, Wang, He, Liu et~al.}]{liu2023mmbench}
Yuan Liu, Haodong Duan, Yuanhan Zhang, Bo~Li, Songyang Zhang, Wangbo Zhao, Yike Yuan, Jiaqi Wang, Conghui He, Ziwei Liu, et~al. 2023.
\newblock Mmbench: Is your multi-modal model an all-around player?
\newblock \emph{arXiv preprint arXiv:2307.06281}.

\bibitem[{Lu et~al.(2023)Lu, Bansal, Xia, Liu, Li, Hajishirzi, Cheng, Chang, Galley, and Gao}]{lu2023mathvista}
Pan Lu, Hritik Bansal, Tony Xia, Jiacheng Liu, Chunyuan Li, Hannaneh Hajishirzi, Hao Cheng, Kai-Wei Chang, Michel Galley, and Jianfeng Gao. 2023.
\newblock Mathvista: Evaluating mathematical reasoning of foundation models in visual contexts.
\newblock \emph{arXiv preprint arXiv:2310.02255}.

\bibitem[{Lu et~al.(2024)Lu, Li, Chen, Xu, Luo, Zhang, and Ye}]{lu2024ovis}
Shiyin Lu, Yang Li, Qing-Guo Chen, Zhao Xu, Weihua Luo, Kaifu Zhang, and Han-Jia Ye. 2024.
\newblock Ovis: Structural embedding alignment for multimodal large language model.
\newblock \emph{arXiv preprint arXiv:2405.20797}.

\bibitem[{Ma{\l}ki{\'n}ski and Ma{\'n}dziuk(2023)}]{malkinski2023review}
Miko{\l}aj Ma{\l}ki{\'n}ski and Jacek Ma{\'n}dziuk. 2023.
\newblock A review of emerging research directions in abstract visual reasoning.
\newblock \emph{Information Fusion}, 91:713--736.

\bibitem[{Masry et~al.(2022)Masry, Long, Tan, Joty, and Hoque}]{masry-etal-2022-chartqa}
Ahmed Masry, Do~Long, Jia~Qing Tan, Shafiq Joty, and Enamul Hoque. 2022.
\newblock \href {https://doi.org/10.18653/v1/2022.findings-acl.177} {{C}hart{QA}: A benchmark for question answering about charts with visual and logical reasoning}.
\newblock In \emph{Findings of the Association for Computational Linguistics: ACL 2022}, pages 2263--2279, Dublin, Ireland. Association for Computational Linguistics.

\bibitem[{Reid et~al.(2024)Reid, Savinov, Teplyashin, Lepikhin, Lillicrap, Alayrac, Soricut, Lazaridou, Firat, Schrittwieser et~al.}]{reid2024gemini}
Machel Reid, Nikolay Savinov, Denis Teplyashin, Dmitry Lepikhin, Timothy Lillicrap, Jean-baptiste Alayrac, Radu Soricut, Angeliki Lazaridou, Orhan Firat, Julian Schrittwieser, et~al. 2024.
\newblock Gemini 1.5: Unlocking multimodal understanding across millions of tokens of context.
\newblock \emph{arXiv preprint arXiv:2403.05530}.

\bibitem[{Sahoo et~al.(2024)Sahoo, Singh, Saha, Jain, Mondal, and Chadha}]{sahoo2024systematic}
Pranab Sahoo, Ayush~Kumar Singh, Sriparna Saha, Vinija Jain, Samrat Mondal, and Aman Chadha. 2024.
\newblock A systematic survey of prompt engineering in large language models: Techniques and applications.
\newblock \emph{arXiv preprint arXiv:2402.07927}.

\bibitem[{Sakaguchi et~al.(2021)Sakaguchi, Bras, Bhagavatula, and Choi}]{sakaguchi2021winogrande}
Keisuke Sakaguchi, Ronan~Le Bras, Chandra Bhagavatula, and Yejin Choi. 2021.
\newblock Winogrande: An adversarial winograd schema challenge at scale.
\newblock \emph{Communications of the ACM}, 64(9):99--106.

\bibitem[{Shao et~al.(2024)Shao, Qian, Xiao, Song, Zong, Wang, Liu, and Li}]{shao2024visual}
Hao Shao, Shengju Qian, Han Xiao, Guanglu Song, Zhuofan Zong, Letian Wang, Yu~Liu, and Hongsheng Li. 2024.
\newblock Visual cot: Unleashing chain-of-thought reasoning in multi-modal language models.
\newblock \emph{arXiv preprint arXiv:2403.16999}.

\bibitem[{Shi et~al.(2023)Shi, Peng, Liao, Lin, Chen, Liu, Zhang, and Jin}]{shi2023exploring}
Yongxin Shi, Dezhi Peng, Wenhui Liao, Zening Lin, Xinhong Chen, Chongyu Liu, Yuyi Zhang, and Lianwen Jin. 2023.
\newblock Exploring ocr capabilities of gpt-4v (ision): A quantitative and in-depth evaluation.
\newblock \emph{arXiv preprint arXiv:2310.16809}.

\bibitem[{Soni et~al.(2022)Soni, Gudala, Pajouhi, and Roberts}]{soni2022radqa}
Sarvesh Soni, Meghana Gudala, Atieh Pajouhi, and Kirk Roberts. 2022.
\newblock \href {https://aclanthology.org/2022.lrec-1.672} {Radqa: A question answering dataset to improve comprehension of radiology reports}.
\newblock In \emph{Proceedings of the Language Resources and Evaluation Conference}, pages 6250--6259, Marseille, France.

\bibitem[{Srivastava et~al.(2022)Srivastava, Rastogi, Rao, Shoeb, Abid, Fisch, Brown, Santoro, Gupta, Garriga-Alonso et~al.}]{srivastava2022beyond}
Aarohi Srivastava, Abhinav Rastogi, Abhishek Rao, Abu Awal~Md Shoeb, Abubakar Abid, Adam Fisch, Adam~R Brown, Adam Santoro, Aditya Gupta, Adri{\`a} Garriga-Alonso, et~al. 2022.
\newblock Beyond the imitation game: Quantifying and extrapolating the capabilities of language models.
\newblock \emph{arXiv preprint arXiv:2206.04615}.

\bibitem[{Stern(1914)}]{stern1914psychological}
William Stern. 1914.
\newblock \emph{The psychological methods of testing intelligence}.
\newblock 13. Warwick \& York.

\bibitem[{Sun et~al.(2024)Sun, Han, Zhao, Ma, Shen, Chen, Chen, and Yu}]{sun2024scieval}
Liangtai Sun, Yang Han, Zihan Zhao, Da~Ma, Zhennan Shen, Baocai Chen, Lu~Chen, and Kai Yu. 2024.
\newblock Scieval: A multi-level large language model evaluation benchmark for scientific research.
\newblock In \emph{Proceedings of the AAAI Conference on Artificial Intelligence}, 17, pages 19053--19061.

\bibitem[{Suzgun et~al.(2022)Suzgun, Scales, Sch{\"a}rli, Gehrmann, Tay, Chung, Chowdhery, Le, Chi, Zhou et~al.}]{suzgun2022challenging}
Mirac Suzgun, Nathan Scales, Nathanael Sch{\"a}rli, Sebastian Gehrmann, Yi~Tay, Hyung~Won Chung, Aakanksha Chowdhery, Quoc~V Le, Ed~H Chi, Denny Zhou, et~al. 2022.
\newblock Challenging big-bench tasks and whether chain-of-thought can solve them.
\newblock \emph{arXiv preprint arXiv:2210.09261}.

\bibitem[{Talmor et~al.(2019)Talmor, Herzig, Lourie, and Berant}]{talmor-etal-2019-commonsenseqa}
Alon Talmor, Jonathan Herzig, Nicholas Lourie, and Jonathan Berant. 2019.
\newblock \href {https://doi.org/10.18653/v1/N19-1421} {{C}ommonsense{QA}: A question answering challenge targeting commonsense knowledge}.
\newblock In \emph{Proceedings of the 2019 Conference of the North {A}merican Chapter of the Association for Computational Linguistics: Human Language Technologies, Volume 1 (Long and Short Papers)}, pages 4149--4158, Minneapolis, Minnesota. Association for Computational Linguistics.

\bibitem[{Team et~al.(2023)Team, Anil, Borgeaud, Wu, Alayrac, Yu, Soricut, Schalkwyk, Dai, Hauth et~al.}]{team2023gemini}
Gemini Team, Rohan Anil, Sebastian Borgeaud, Yonghui Wu, Jean-Baptiste Alayrac, Jiahui Yu, Radu Soricut, Johan Schalkwyk, Andrew~M Dai, Anja Hauth, et~al. 2023.
\newblock Gemini: a family of highly capable multimodal models.
\newblock \emph{arXiv preprint arXiv:2312.11805}.

\bibitem[{Thrush et~al.(2022)Thrush, Jiang, Bartolo, Singh, Williams, Kiela, and Ross}]{thrush2022winoground}
Tristan Thrush, Ryan Jiang, Max Bartolo, Amanpreet Singh, Adina Williams, Douwe Kiela, and Candace Ross. 2022.
\newblock Winoground: Probing vision and language models for visio-linguistic compositionality.
\newblock In \emph{Proceedings of the IEEE/CVF Conference on Computer Vision and Pattern Recognition}, pages 5238--5248.

\bibitem[{Touvron et~al.(2023)Touvron, Lavril, Izacard, Martinet, Lachaux, Lacroix, Rozi{\`e}re, Goyal, Hambro, Azhar et~al.}]{touvron2023llama}
Hugo Touvron, Thibaut Lavril, Gautier Izacard, Xavier Martinet, Marie-Anne Lachaux, Timoth{\'e}e Lacroix, Baptiste Rozi{\`e}re, Naman Goyal, Eric Hambro, Faisal Azhar, et~al. 2023.
\newblock Llama: Open and efficient foundation language models.
\newblock \emph{arXiv preprint arXiv:2302.13971}.

\bibitem[{Wang et~al.(2023{\natexlab{a}})Wang, Lv, Yu, Hong, Qi, Wang, Ji, Yang, Zhao, Song, Xu, Xu, Li, Dong, Ding, and Tang}]{wang2023cogvlm}
Weihan Wang, Qingsong Lv, Wenmeng Yu, Wenyi Hong, Ji~Qi, Yan Wang, Junhui Ji, Zhuoyi Yang, Lei Zhao, Xixuan Song, Jiazheng Xu, Bin Xu, Juanzi Li, Yuxiao Dong, Ming Ding, and Jie Tang. 2023{\natexlab{a}}.
\newblock \href {http://arxiv.org/abs/2311.03079} {Cogvlm: Visual expert for pretrained language models}.

\bibitem[{Wang et~al.(2023{\natexlab{b}})Wang, Hu, Lu, Zhu, Zhang, Subramaniam, Loomba, Zhang, Sun, and Wang}]{wang2023scibench}
Xiaoxuan Wang, Ziniu Hu, Pan Lu, Yanqiao Zhu, Jieyu Zhang, Satyen Subramaniam, Arjun~R Loomba, Shichang Zhang, Yizhou Sun, and Wei Wang. 2023{\natexlab{b}}.
\newblock Scibench: Evaluating college-level scientific problem-solving abilities of large language models.
\newblock \emph{arXiv preprint arXiv:2307.10635}.

\bibitem[{Wang and Chiew(2010)}]{wang2010cognitive}
Yingxu Wang and Vincent Chiew. 2010.
\newblock On the cognitive process of human problem solving.
\newblock \emph{Cognitive systems research}, 11(1):81--92.

\bibitem[{Wang et~al.(2024)Wang, Chen, Han, Lin, Zhao, Liu, Zhai, Yuan, You, and Yang}]{wang2024exploring}
Yiqi Wang, Wentao Chen, Xiaotian Han, Xudong Lin, Haiteng Zhao, Yongfei Liu, Bohan Zhai, Jianbo Yuan, Quanzeng You, and Hongxia Yang. 2024.
\newblock Exploring the reasoning abilities of multimodal large language models (mllms): A comprehensive survey on emerging trends in multimodal reasoning.
\newblock \emph{arXiv preprint arXiv:2401.06805}.

\bibitem[{Wei et~al.(2022)Wei, Wang, Schuurmans, Bosma, Xia, Chi, Le, Zhou et~al.}]{wei2022chain}
Jason Wei, Xuezhi Wang, Dale Schuurmans, Maarten Bosma, Fei Xia, Ed~Chi, Quoc~V Le, Denny Zhou, et~al. 2022.
\newblock Chain-of-thought prompting elicits reasoning in large language models.
\newblock \emph{Advances in neural information processing systems}, 35:24824--24837.

\bibitem[{Xu et~al.(2024)Xu, Wang, Liu, and Xu}]{xu2024collage}
Siyu Xu, Yunke Wang, Daochang Liu, and Chang Xu. 2024.
\newblock Collage prompting: Budget-friendly visual recognition with gpt-4v.
\newblock \emph{arXiv preprint arXiv:2403.11468}.

\bibitem[{Yue et~al.(2024)Yue, Ni, Zhang, Zheng, Liu, Zhang, Stevens, Jiang, Ren, Sun et~al.}]{yue2024mmmu}
Xiang Yue, Yuansheng Ni, Kai Zhang, Tianyu Zheng, Ruoqi Liu, Ge~Zhang, Samuel Stevens, Dongfu Jiang, Weiming Ren, Yuxuan Sun, et~al. 2024.
\newblock Mmmu: A massive multi-discipline multimodal understanding and reasoning benchmark for expert agi.
\newblock In \emph{Proceedings of the IEEE/CVF Conference on Computer Vision and Pattern Recognition}, pages 9556--9567.

\bibitem[{Zhang et~al.(2019)Zhang, Gao, Jia, Zhu, and Zhu}]{zhang2019raven}
Chi Zhang, Feng Gao, Baoxiong Jia, Yixin Zhu, and Song-Chun Zhu. 2019.
\newblock Raven: A dataset for relational and analogical visual reasoning.
\newblock In \emph{Proceedings of the IEEE/CVF conference on computer vision and pattern recognition}, pages 5317--5327.

\bibitem[{Zhang et~al.(2024{\natexlab{a}})Zhang, Jiang, Zhang, Lin, Guo, Qiu, Zhou, Lu, Chang, Gao et~al.}]{zhang2024mathverse}
Renrui Zhang, Dongzhi Jiang, Yichi Zhang, Haokun Lin, Ziyu Guo, Pengshuo Qiu, Aojun Zhou, Pan Lu, Kai-Wei Chang, Peng Gao, et~al. 2024{\natexlab{a}}.
\newblock Mathverse: Does your multi-modal llm truly see the diagrams in visual math problems?
\newblock \emph{arXiv preprint arXiv:2403.14624}.

\bibitem[{Zhang et~al.(2024{\natexlab{b}})Zhang, Aljunied, Gao, Chia, and Bing}]{zhang2024m3exam}
Wenxuan Zhang, Mahani Aljunied, Chang Gao, Yew~Ken Chia, and Lidong Bing. 2024{\natexlab{b}}.
\newblock M3exam: A multilingual, multimodal, multilevel benchmark for examining large language models.
\newblock \emph{Advances in Neural Information Processing Systems}, 36.

\bibitem[{Zhang et~al.(2023)Zhang, Zhang, Li, Zhao, Karypis, and Smola}]{zhang2023multimodal}
Zhuosheng Zhang, Aston Zhang, Mu~Li, Hai Zhao, George Karypis, and Alex Smola. 2023.
\newblock Multimodal chain-of-thought reasoning in language models.
\newblock \emph{arXiv preprint arXiv:2302.00923}.

\bibitem[{Zhao et~al.(2023)Zhao, Quan, Zhu, and Yang}]{zhao2023clip4str}
Shuai Zhao, Ruijie Quan, Linchao Zhu, and Yi~Yang. 2023.
\newblock Clip4str: A simple baseline for scene text recognition with pre-trained vision-language model.
\newblock \emph{arXiv preprint arXiv:2305.14014}.

\end{thebibliography}
\bibliographystyle{acl_natbib}
\newpage
\appendix

\section{Appendix}
\label{sec:appendix}

\subsection{Additional Dataset Details}\label{ref:datasetDetails}

Table \ref{tab:dataset_cat_count},  reveals a skewed distribution across various question categories. Notably, textual categories like the Alphabet Test (ALP) and Mathematical Operations (MTO) contain 94 and 99 examples, respectively. In contrast, many vision-based categories are more challenging and typically include between 80 and 100 examples. For instance, the Non-Verbal Analogy (NVA) category, one of the most difficult, comprises 100 examples. Although this skewed distribution could impact model performance, exploring its effects is beyond the scope of this manuscript and is left for future work.

\noindent\textbf{Problem Sub categories.} The above categories are further subdivided based on the different modality of input type(text or text-vision). Table \ref{tab:dataset_cat_count} lists each sub-category for \textbf{Text Only} questions and \textbf{Text+Vision} questions, along with the respective count for each category. As in Table \ref{tab:dataset_cat_count} most categories are represented well in the dataset. 

\paragraph{Problem Categories}
Problem categories in the dataset with description are shown in Table \ref{tab:category_desc}.
\label{sec:porblem_categories}

\begin{table}[!htb]
\scriptsize
\setlength{\tabcolsep}{3.5pt}
\centering
\begin{tabular}{@{}p{1.8cm}|p{5.0cm}@{}}
\textbf{Category} & \textbf{Description}\\
\hline
Series & Finding the missing element in numerical, alphabetical, or alphanumeric series. \\
\hline
Alphabet Test & Focusing on operations involving the English alphabet, such as anagrams. \\
\hline
Classification/Odd One Out & Identifying the item that is different from the others. \\
\hline
Analogy & Solving problems of the type a:b::c:? \\
\hline
Coding-Decoding & Deciphering codes and symbols to infer rules and apply them to new examples. \\
\hline
Number and Ranking & Calculating occurrences or determining order based on certain properties. \\
\hline
Blood Relation & Solving problems based on family relationships. \\
\hline
Mathematical Operations & Using mathematical operations like addition, multiplication, subtraction, and division to solve problems. \\
\hline
Direction Sense & Determining direction and location based on given instructions. \\
\hline
Venn Diagrams & Using set theory and relationships depicted in Venn diagrams to solve problems. \\
\hline
Time and Clock & Calculating dates, days, and times based on given information. \\
\hline
Missing Character & Predicting the missing element in a figure, requiring spatial thinking. \\
\hline
Non-Verbal Series & Predicting the next element in a sequence of figures. \\
\hline
Non-Verbal Classification/Non-Verbal Odd One Out & Identifying the figure that is different from the others. \\
\hline
Non-Verbal Analogy & Solving analogy problems of the type a:b::c:? using figures. \\
\hline
Incomplete Figure & Identifying the missing part of a figure. \\
\hline
Mirror, Water and Images & Solving problems related to reflections and image transformations. \\
\hline
Cube and Dice & Solving problems involving painting, counting, and manipulating cubes and dice. \\
\hline
Paper Folding \& Cutting & Determining the resulting shape after paper folding and cutting. \\
\hline
Embedded Figure & Finding the alternative which contains a given figure as its part. \\
\hline
Puzzle Test & Solving general puzzles involving arrangement and deduction. \\
\hline
Figure Partition & Calculating the number of specific shapes (like triangles) in a figure. \\
\hline
Dot Problem & Finding similar conditions in alternative figures based on dot arrangements. \\
\hline
Cryptography & Deciphering codes that involve arithmetic operations. \\
\hline
Syllogisms & Making inferences based on given statements, often solved using Venn diagrams. \\
\hline
Statement \& Conclusions & Making inferences based on given statements, not typically solved with Venn diagrams. \\
\hline
Data Sufficiency & Determining whether the given information is sufficient to solve a problem. 
\end{tabular}
\caption{\textbf{\datasetName Problem Categories}. This table lists broad categories of problems that frequently appear in the NTSE exam. \textit{Note: This is not an exhaustive list; other types of questions may also appear in the NTSE exam.}}
\label{tab:category_desc}
\vspace{-1.5em}
\end{table}

\subsubsection{Data Extraction Pipeline}\label{ref:appendix:ExtractionPipeline}
The questions are extracted from the sources listed in section \ref{section:dataset_sources}, with human intervention to monitor and rectify mistakes made by the automated pipeline. 
The data extraction pipeline involves first processing the PDF through {\sc MathPix OCR}\footnote{\href{https://mathpix.com/ocr}{MathPix OCR}} to generate a Word file, which was then manually corrected for any errors. Next, we used the {\sc docxlatex} \footnote{\href{https://pypi.org/project/docxlatex/}{docxlatex}} library to convert all equations into LaTeX expressions. Finally, we leveraged the {\sc PyMuPDF} \footnote{\href{https://pymupdf.readthedocs.io/en/latest/}{PyMuPDF}} library to extract all text and images, extracting :
\begin{itemize}
    \item \textbf{(1) Textual data} i.e. direction (extra guidance on the context of the question), the question, the correct answer option and the solution. Any errors in text extraction was rectified by human annotator. 
    \item \textbf{(2) Vision Based data} i.e. relevant image/s which we segregate into direction images, problem images, option images, and solution images. Questions associated with low-quality images were excluded. A human annotator assisted in identifying these low-quality images.
\end{itemize}

A total of 2,728 MCQ (multiple-choice questions) consisting of a total of 4,642 images across 26 categories of questions are extracted and \datasetName is created along with the necessary metadata. 

The correct answers to the questions are also officially provided by the NTSE exam organizers (under NCERT). The exam is renowned for its high quality, as the questions are typically designed by subject-matter experts. Any disputes or challenges regarding incorrect answers or flawed questions are meticulously reviewed by NCERT before the release of the final answer key, ensuring the accuracy and reliability of the solutions, ensuring good quality data. 

\subsubsection{Dataset Sources Links}
\label{section:dataset_sources_links}
(a.) Resonance \footnote{\href{https://www.resonance.ac.in/answer-key-solutions/ntse-stage-I.aspx}{Paper Links}}. 
(b.) Reference book titled \textit{A Modern Approach to Verbal and Non-Verbal Reasoning} \footnote{\href{https://www.schandpublishing.com/books/competitive-books/dr-rs-aggarwal/a-modern-approach-verbal-non-verbal-reasoning/9789355011534/}{A Modern Approach to Verbal and Non-Verbal Reasoning}}. 
(c.) Another book titled \textit{Study Guide for NTSE} \footnote{\href{https://dishapublication.com/products/mega-study-guide-for-ntse-sat-mat-class-10-stage-1-2-paperback-disha-experts?_pos=1&_sid=40b3fe6bf&_ss=r}{Study Guide for NTSE}}.

\subsection{Error Analysis for different question categories}
\label{sec:appx:reason_analysis}
Analyzing the incorrect responses of Gemini 1.5 Pro across different categories highlighted several error patterns and identified key areas for improvement.

\textbf{Alphabet Test:} Common errors involve incorrect word ordering and miscalculated letter positions, resulting in faulty alphabetical arrangements. The model needs improved techniques for letter manipulation and more accurate application of alphabetical rules.

\textbf{Analogy:} The model struggles with identifying underlying relationships in both verbal and non-verbal analogies, often misinterpreting visual patterns. Enhancing pattern recognition, especially in visual contexts, would strengthen performance.

\textbf{Blood Relation:} A frequent issue is the misinterpretation of family relationships, particularly when mapping out complex family trees. Focused improvements in logical reasoning around relational structures can address this.

\textbf{Odd One Out:} Errors often arise from misidentifying the unique element in a set, as the model fails to consistently recognize distinguishing patterns. Better classification based on subtle attribute differences is needed.

\textbf{Coding-Decoding:} Incorrect application of coding schemes, such as letter shifts or reversals, is common. The model also fails to detect important details in coding patterns, highlighting the need for more attention to detail.

\textbf{Cube and Dice:} The model exhibits difficulty with visualizing 3D objects and their geometric properties, leading to errors in counting cube faces or determining surface areas. Spatial reasoning should be strengthened in these cases.

\textbf{Data Sufficiency:} Misinterpretation of the sufficiency of provided statements to solve a problem is a frequent error. The model often fails to draw correct conclusions from quantitative data, indicating a need for better assessment and logic application.

\textbf{Direction Sense:} The model often misinterprets movements and directions, leading to incorrect final positions. Quantitative errors also arise in calculating distances or angles, suggesting that spatial reasoning and movement tracking must be enhanced.

\textbf{Embedded Figure:} The model has difficulty recognizing shapes embedded within larger figures, especially when they are rotated. This calls for improved shape recognition regardless of orientation.

\textbf{Figure Partition:} Errors in counting the components of complex figures, such as lines or triangles, often occur due to difficulty visualizing accurate partitioning. Improved counting strategies and partitioning techniques are needed.

\textbf{Incomplete Figure:} The model misidentifies the missing part of a figure due to incorrect application of rotation or pattern recognition. Enhancing geometric transformation abilities and pattern identification can mitigate this.

\textbf{Missing Character:} The model struggles with recognizing underlying alphabetical or numerical patterns and sometimes makes calculation errors after identifying the pattern. Strengthening both sequence recognition and calculation accuracy is important.

\textbf{Non-Verbal Analogy:} Visual analogy problems reveal difficulties in identifying figure transformations such as rotations or reflections. The model would benefit from improved spatial reasoning and visual transformation handling.

\textbf{Non-Verbal Odd One Out:} The model frequently misidentifies the attribute that sets the odd figure apart in a visual set. Enhancing its ability to classify subtle visual differences can improve performance.

\textbf{Number and Ranking:} Errors in determining sequences or ranks in numerical problems arise from miscalculations. A stronger grasp of ranking principles and numerical reasoning is essential for more accurate outcomes.

\textbf{Paper Folding \& Cutting:} The model has difficulty predicting the results of folds and cuts, often misinterpreting how patterns will replicate upon unfolding. 

\textbf{Puzzle Test:} Incorrect interpretation of puzzle clues or failure to apply logical deduction correctly are common errors. The model also struggles with numerical-based puzzle elements, suggesting a need for more systematic problem-solving approaches.

\textbf{Series:} The model often fails to recognize numerical or alphabetical progression patterns and makes calculation errors after identifying a pattern. Improved pattern detection and calculation accuracy would enhance performance.

\textbf{Statement \& Conclusions:} Misapplications of logical deduction, particularly in determining whether a conclusion follows from a given statement, are frequent errors. Reinforcing critical thinking skills and logical structure comprehension will help address these issues.

\textbf{Syllogisms:} The model struggles with applying syllogistic rules correctly and often misinterprets categorical relationships. A stronger understanding of syllogistic reasoning and relational logic is necessary.

\textbf{Time and Clock:} Miscalculations related to time intervals or angles, particularly involving exceptions, often lead to errors. Enhancing the model’s mathematical reasoning around time and angles will improve performance in these questions.

\textbf{Venn Diagrams:} Common errors include misrepresentation of set relationships and logical errors in diagram interpretation. Enhancing understanding of set theory and logical interpretation will help reduce errors.

\textbf{Mathematical Operations:} The model often misapplies sign changes or substitutions, and errors occur when performing calculations after applying these changes. Focusing on improving accuracy in mathematical operations and sign transformations is necessary.

\textbf{Mirror, Water, \& Images:} Incorrect application of reflection principles leads to selection errors, and the model struggles with predicting figure changes under reflection. Strengthening spatial visualization and understanding reflection principles will improve outcomes.

\textbf{Non-Verbal Series:} The model often misinterprets the progression of visual figure transformations, leading to incorrect sequence predictions. Enhancing pattern recognition and spatial reasoning for visual sequences is needed.

\textbf{Dot Problem:} The model fails to correctly identify overlapping regions for dot placement and struggles with understanding the spatial relationships between shapes. Better understanding of intersections and spatial overlap is crucial.
\subsection{Additional Results}
\subsubsection{Preliminary Human Evaluation}\label{sec:appx:human_eval}
Two human annotators were tasked with solving 10 questions for each category for both textual and multi-modality questions types. Results are shown in Table \ref{tab:human_eval_text} and \ref{tab:human_eval_multimodal}. Average accuracy for textual category is close to 85\%, and for multi-modal category is around 83.33\%. 

\begin{table*}[!htb]
\setlength{\tabcolsep}{2pt} 
\small
\centering
\begin{tabular}{l|cccccccccccc|c}
\hline
Annotator & SER & ALP & ODO & ANA & COD & NUM & BLR & MTO & PUZ & SYL & STC & DAT & Aggregate \\ 
\hline
\textit{First} & 90	& 90 & 80 &	100	& 90	& 90	& 80 &	90 &	100 &	80 &	100 &	80 &	89.17 \\
\textit{Second} & 80 &	90 &	90 & 	80	& 80 & 	80 &	90 &	100 &	90 &	60 &	80 &	70 &	82.50 \\
\textit{Third} & 80	& 80	& 90	& 90 &	90 &	90 &	80 &	90 &	90 &	70 &	80 &	70 &	83.33\\
\hline
\end{tabular}
\caption{\textbf{Human Evaluation of Textual Questions.} Percentage accuracy for 3 human annotators.}
\label{tab:human_eval_text}
\end{table*}

\begin{table*}[!htb]
\setlength{\tabcolsep}{2.0pt} 
\centering
\small

\begin{tabular}{l|cccccccccccccc|c}
\hline
Annotator & DIR & VEN & TIM & MIS & NVS & NVO & NVA & INC & MIR & CUB & PAP & EMB & FIG & DOT & Aggregate \\ \hline
\textit{First} & 100 & 100 & 60 & 80 & 80 & 60 & 90 & 100 & 80 & 90 & 90& 100 & 70 & 100 & 85.71 \\
\textit{Second} & 90 & 70 & 50 & 100 & 100 & 50 & 100 & 90 & 70 & 90 & 70 & 90 & 90 & 60 & 80.71 \\
\textit{Third} & 80 &	90 &	70 & 	100 &	70 &	50 &	70 &	90 &	100 &	80 &	80 &	100 &	90 &	100 &	83.57\\
\hline
\end{tabular}
\caption{\textbf{Human Evaluation of Multi-modal Questions.} Percentage accuracy for 2 human annotators.}
\label{tab:human_eval_multimodal}
\end{table*}

\subsubsection{Option Shuffling Experiment Results}
Results for textual questions for Gemini 1.5 Pro and GPT-4o are shown in Tables ~\ref{tab:option_shuff_gemini_text} and ~\ref{tab:option_shuff_gpt_text}. Results for multi-modal questions for Gemini 1.5 Pro and GPT-4o are shown in Tables ~\ref{tab:option_shuff_gpt_vis} and ~\ref{tab:option_shuff_gemini_vis}. 

\paragraph{Additional Analysis of Option Shuffling Experiment}: The performance of random set is close to the original set for GPT-4o, but has a significant difference(4-6 percent) for Gemini 1.5 Pro model, for both textual and multimodel questions. This result suggests that Gemini1.5 Pro may be more prone to memorizing results compared to GPT-4o, as it shows greater sensitivity to the position of the options.

\begin{table*}[!htb]
\setlength{\tabcolsep}{2pt} 
\centering
\begin{tabular}{l|cccccccccccc|c}
\hline
Ans Option & SER & ALP & ODO & ANA & COD & NUM & BLR & MTO & PUZ & SYL & STC & DAT & Aggregate \\ \hline
\textit{First} & 59.90 & 21.99 & 53.92 & 62.69 & 41.61 & 60.19 & 59.26 & 37.37 & 70.18 & 50.00 & 67.95 & 49.26 & 52.86 \\
\textit{Second} & 65.75 & 26.24 & 60.59 & 69.09 & 46.76 & 66.67 & 67.72 & 43.09 & 71.23 & 63.64 & 66.02 & 46.67 & 57.79 \\
\textit{Third} & 61.72 & 56.03 & 62.94 & 73.07 & 51.68 & 71.46 & 67.46 & 42.08 & 70.18 & 53.03 & 69.87 & 56.30 & 61.32 \\
\textit{Fourth} & 62.24 & 28.02 & 58.82 & 67.55 & 48.77 & 72.18 & 68.78 & 51.85 & 71.93 & 47.73 & 70.83 & 49.26 & 58.16 \\
Random & 62.11 & 25.18 & 57.26 & 68.87 & 53.02 & 62.83 & 68.25 & 47.81 & 74.03 & 47.73 & 69.23 & 45.19 & 56.79 \\ 
Original & 63.67 &39.36 &60.00 &69.54 &61.07 &68.35 &58.73 &45.45 &81.05 &65.91 &70.19 &63.33 &62.22 \\
\hline
\end{tabular}
\caption{\textbf{Gemini 1.5 Pro Option Shuffling Results on Text Questions.} \textit{Original,} is the option arrangement same as in the original question paper. \textit{Random,} is random position of the correct option.\textit{First, Second, Third, Fourth} are the position of correct answers.}
\label{tab:option_shuff_gemini_text}
\end{table*}

\begin{table*}[!htb]
\setlength{\tabcolsep}{2.0pt} 
\centering
\scalebox{0.88}{
\begin{tabular}{l|cccccccccccccc|c}
\hline
Ans Option & DIR & VEN & TIM & MIS & NVS & NVO & NVA & INC & MIR & CUB & PAP & EMB & FIG & DOT & Aggregate \\ \hline
\textit{First} & 57.29 & 50.75 & 56.86 & 34.91 & 50.88 & 34.29 & 40.00 & 54.26 & 52.90 & 42.32 & 28.82 & 22.92 & 45.54 & 33.33 & 43.22 \\
\textit{Second} & 52.43 & 63.96 & 64.71 & 34.12 & 31.23 & 23.81 & 41.33 & 26.24 & 27.54 & 38.58 & 37.85 & 31.60 & 32.39 & 23.19 & 37.78 \\
\textit{Third} & 67.36 & 63.06 & 68.63 & 48.55 & 20.00 & 20.48 & 22.00 & 24.82 & 22.46 & 40.45 & 21.88 & 27.08 & 40.38 & 40.58 & 37.70 \\
\textit{Fourth} & 59.03 & 59.46 & 63.40 & 35.70 & 9.12 & 11.43 & 18.00 & 15.25 & 32.97 & 48.69 & 27.78 & 24.30 & 44.60 & 10.15 & 32.85 \\
Random & 62.15 & 58.86 & 58.82 & 33.86 & 22.46 & 23.33 & 34.00 & 30.50 & 29.35 & 46.07 & 34.38 & 26.39 & 38.50 & 27.54 & 37.59 \\ 
Original & 63.54 & 64.86 & 70.59 & 37.01 & 33.68 & 25.71 & 32 & 38.3 & 35.87 & 43.82 & 30.21 & 36.46 & 46.48 & 30.43 & 42.06  \\ \hline
\end{tabular}
}
\caption{\textbf{Gemini 1.5 Pro Option Shuffling Results on Multi-modal Questions.} \textit{Original,} is the option arrangement same as in the original question paper. \textit{Random,} is random position of the correct option. \textit{First, Second, Third, Fourth} are the position of correct answers.}
\label{tab:option_shuff_gemini_vis}
\end{table*}

\begin{table*}[!htb]
\setlength{\tabcolsep}{2pt} 
\centering
\begin{tabular}{l|cccccccccccc|c}
\hline
Ans Option & SER & ALP & ODO & ANA & COD & NUM & BLR & MTO & PUZ & SYL & STC & DAT & Aggregate\\ \hline
\textit{First} & 52.34 & 29.79 & 47.45 & 64.46 & 50.34 & 44.60 & 65.08 & 36.70 & 64.56 & 57.58 & 70.19 & 51.48 & 52.88\\
\textit{Second} & 45.96 & 37.94 & 64.12 & 60.93 & 47.88 & 46.28 & 66.40 & 38.72 & 65.62 & 62.12 & 73.72 & 51.11 & 55.07\\
\textit{Third} & 42.32 & 38.65 & 63.33 & 60.26 & 42.73 & 44.36 & 64.55 & 38.04 & 66.67 & 57.58 & 72.44 & 48.15 & 53.26\\
\textit{Fourth} & 34.38 & 39.00 & 59.41 & 56.07 & 35.57 & 39.81 & 60.84 & 40.40 & 67.72 & 65.15 & 75.00 & 55.93 & 52.44\\
Random & 44.53 & 35.46 & 56.67 & 61.15 & 46.09 & 41.25 & 64.02 & 35.69 & 66.67 & 60.61 & 75.00 & 48.15 & 52.94\\ 
Original &42.58 &35.11 &55.88 &65.56 &38.26 &42.45 &68.25 &41.41 &69.47 &63.64 &70.19 &43.33 &53.01 \\
\hline
\end{tabular}
\caption{\textbf{GPT-4o Option Shuffling Results on Textual Questions.} \textit{Original,} is the option arrangement same as in the original question paper. \textit{Random,} is random position of the correct option. \textit{First, Second, Third, Fourth} are the position of correct answers.}
\label{tab:option_shuff_gpt_text}
\end{table*}

\begin{table*}[!htb]
\setlength{\tabcolsep}{2.0pt} 
\centering
\scalebox{0.88}{
\begin{tabular}{l|cccccccccccccc|c}
\hline
Ans Option & DIR & VEN & TIM & MIS & NVS & NVO & NVA & INC & MIR & CUB & PAP & EMB & FIG & DOT & Aggregate \\ \hline
\textit{First} & 44.10 & 61.56 & 31.37 & 29.13 & 23.86 & 32.86 & 24.67 & 12.41 & 30.07 & 41.20 & 31.25 & 21.88 & 52.11 & 8.70 & 31.80 \\
\textit{Second} & 42.36 & 54.65 & 43.79 & 32.28 & 22.11 & 29.05 & 31.00 & 29.08 & 40.94 & 49.81 & 42.01 & 26.04 & 46.48 & 30.43 & 37.15 \\
\textit{Third} & 50.00 & 54.95 & 39.87 & 31.23 & 20.70 & 25.24 & 30.33 & 27.31 & 34.42 & 40.08 & 37.85 & 30.91 & 50.24 & 31.88 & 36.07 \\
\textit{Fourth} & 51.73 & 48.35 & 37.25 & 26.25 & 23.86 & 18.57 & 30.67 & 30.14 & 27.17 & 37.45 & 38.54 & 25.35 & 41.31 & 36.23 & 33.78 \\
Random & 44.79 & 54.35 & 37.25 & 29.66 & 22.11 & 23.81 & 24.67 & 25.53 & 32.61 & 38.21 & 35.07 & 30.21 & 39.44 & 23.19 & 32.92 \\ 

Original &37.5 &50.45 &41.18 &29.92 &16.84 &22.86 &26 &23.4 &34.78 &35.96 &27.08 &22.92 &45.07 &17.39 &30.81 \\
\hline
\end{tabular}}
\caption{\textbf{GPT-4o Option Shuffling Results on Multi-modal Questions.} \textit{Original,} is the option arrangement same as in the original question paper. \textit{Random,} is random position of the correct option. \textit{First, Second, Third, Fourth} are the position of correct answers.}
\label{tab:option_shuff_gpt_vis}
\end{table*}

\begin{table*}[!htb]
\setlength{\tabcolsep}{2pt} 
\centering
\begin{tabular}{l|cccccccccccc|c}
\hline
Ans Option & SER & ALP & ODO & ANA & COD & NUM & BLR & MTO & PUZ & SYL & STC & DAT & Aggregate \\ \hline
\textit{First} & 80.63 & 88.15 & 77.65 & 71.78 & 84.09 & 85.51 & 83.33 & 84.69 & 87.80 & 81.08 & 74.04 & 80.95 & 80.88 \\
\textit{Second} & 76.68 & 84.78 & 78.82 & 76.55 & 85.23 & 86.23 & 84.92 & 82.65 & 84.15 & 78.38 & 71.15 & 82.14 & 80.45 \\
\textit{Third} & 83.00 & 84.78 & 83.53 & 79.31 & 86.36 & 83.33 & 84.13 & 79.59 & 86.59 & 78.38 & 75.96 & 82.14 & 82.43 \\
\textit{Fourth} & 81.42 & 85.87 & 78.82 & 77.93 & 86.36 & 86.96 & 83.33 & 82.65 & 89.02 & 75.68 & 79.81 & 80.95 & 82.29 \\ 
\hline
\end{tabular}
\caption{\textbf{O1-preview Option Shuffling Results on Textual Questions.} \textit{First, Second, Third, Fourth} are the position of correct answers. }
\label{tab:option_shuff_o1_text}
\end{table*}

\subsection{Model Details Hyper parameters}\label{ref:appx:model_hyper}
We use the following models for running experiments:

\textbf{LLMs}: GPT-3.5-Turbo, Llama3-70b \cite{llama3modelcard}, Mixtral8x7b \cite{jiang2024mixtral} using the Standard QA strategy with both zero shot COT and few shot COT.

\textbf{Open-Source VLMs}\footnote{Evaluated on A6000 machine}: QWEN-VL-chat-7b \cite{bai2023qwen}, CogVLM-2-Llama3-chat-19B \cite{wang2023cogvlm}, internlm-xcomposer2-vl-7b \cite{dong2024internlm},Ovis1.6-Gemma2-9B\cite{lu2024ovis},LLaVA-OneVision-Qwen2-72b-ov-chat \cite{li2024llava} using the Standard VQA and Image-Only strategies with zero shot COT. \footnote{Refer to technical discussion section for details on why few shot COT is challenging with open source models.}

\textbf{Proprietary VLMs}: Gemini-1.5-Pro \cite{reid2024gemini}, Gemini-1.5-Flash and GPT-4o  are the proprietary models used.
We provide the prompts and hyperparameters in the Appendix \ref{sec:prompt_templates} and \ref{ref:appx:model_hyper}. We have also evaluated the cheaper and faster version of Gemini, namely Gemini-1.5-Flash, which has shown comparable performance to Gemini-1.5-Pro.

Default hyperparameters from the Hugging Face model were used. List along with modifications(if any) are listed below
\begin{itemize}
\item \textbf{GPT-4o.} Temparature = 0.0, Output\_format = json
\item \textbf{GPT-3.5.} Temparature = 0.0, Output\_format = json
\item \textbf{Gemini-1.5-Pro.} Temparature = 1.0
\item \textbf{Qwen-vl-chat.} seed = 1234, precision = float16(half)
\item \textbf{Cogagent2-Llama3-8b.} precision=bf16
\item \textbf{InterLM-XComposer.} precision=half
\item \textbf{LLaVA-OneVision-Qwen2-72b-ov-chat}
\item \textbf{Ovis1.6-Gemma2-9B} 

\end{itemize}

\subsection{Prompt Templates}\label{sec:prompt_templates}

\begin{figure*}
    \centering
    \includegraphics[scale=0.45]{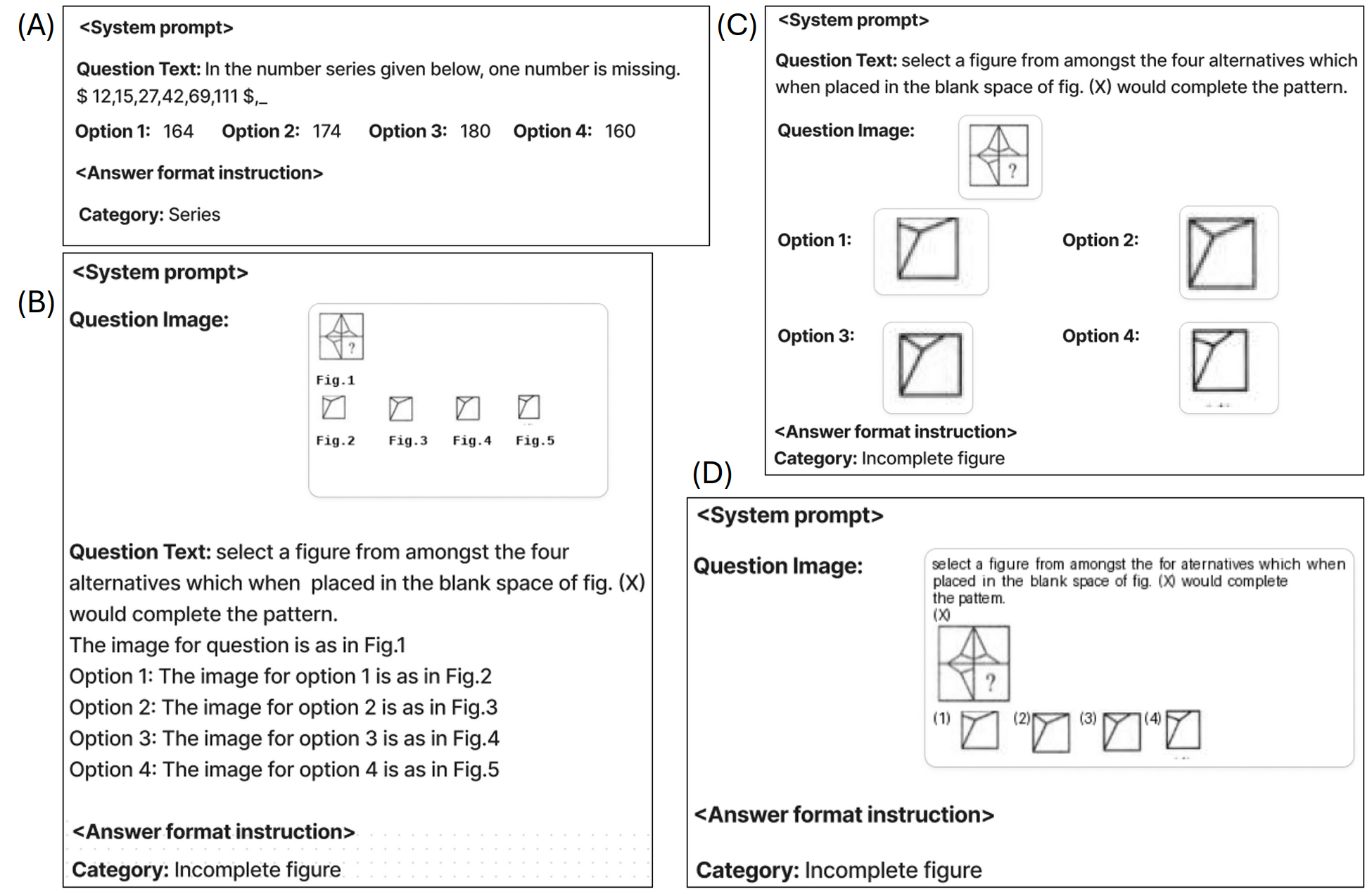}
    \caption{\textbf{Examples Showing Input to Different Proposed modelling strategies}.(A) Text Only Standard QA strategy(B) Standard VQA (C) Interleaved Strategy (D) Image Only.}
    \label{fig:system_prompt}
\end{figure*}

System prompts for different modelling strategies, i.e., \textbf{\textit{Standard QA, Image only, Interleaved and Standard VQA}}, are shown in Figure \ref{fig:system_prompt}. 

We present the prompt templates used across different modeling and prompting strategies for GPT-4o. We apply the same prompt template consistently for each model within the same strategy.

\subsubsection{Interleaving}
\textbf{Zero shot}
\centering

\lstdefinelanguage{json}{
    basicstyle=\small\ttfamily,
    commentstyle=\color{gray},
    showstringspaces=false,
    breaklines=true,
    frame=lines,
    backgroundcolor=\color{white}
    literate=
     *{0}{{{\color{blue}0}}}{1}
      {1}{{{\color{blue}1}}}{1}
      {2}{{{\color{blue}2}}}{1}
      {3}{{{\color{blue}3}}}{1}
      {4}{{{\color{blue}4}}}{1}
      {5}{{{\color{blue}5}}}{1}
      {6}{{{\color{blue}6}}}{1}
      {7}{{{\color{blue}7}}}{1}
      {8}{{{\color{blue}8}}}{1}
      {9}{{{\color{blue}9}}}{1}
      {:}{{{\color{purple}{:}}}}{1}
      {,}{{{\color{purple}{,}}}}{1}
      {\{}{{{\color{brown}{\{}}}}{1}
      {\}}{{{\color{brown}{\}}}}}{1}
      {[}{{{\color{brown}{[}}}}{1}
      {]}{{{\color{brown}{]}}}}{1},
}
\begin{lstlisting}[language=json, caption={Prompt template for GPT-4o zero shot with interleaving. The template includes placeholders (`...`) for question direction, question text, answer choices, and their corresponding images, which are encoded in base64 format. All the images are optional and can be there or not based on the question.}]
{
  "role": "system",
  "content": "You are a brilliant problem solver... Please select the correct answer choice."
},
{
  "role": "user",
  "content": [
    {"type": "text", "text": "questionDirection: ..."},
    {"type": "image_url", "image_url": {"url": "data:image/png;base64,..."}}, 
    {"type": "text", "text": "\nquestionText: ..."},
    {"type": "image_url", "image_url": {"url": "data:image/png;base64,..."}},
    {"type": "text", "text": "\nanswerChoices: \n 1. ... \n 2. ... \noptionImages: \n..."},
    {"type": "text", "text": "Answer in the json format as follows: \n {'answer': <correct_option_number>, 'explanation': <explanation>}"}
  ]
}
\end{lstlisting}

\textbf{Few shot}
\centering
\begin{lstlisting}[language=json, breaklines=true, caption={Few-shot CoT prompt template for GPT-4o.  The template includes placeholders for multiple solved examples (with directions, questions, images, answer choices, correct answers, and explanations) followed by a new unsolved question (with directions, questions, images, and answer choices). GPT-4o is expected to provide a structured answer in JSON format. All the images are optional and can be there or not based on the question.}]
{
  "role": "system",
  "content": "Understand the following problems carefully... then answer the new question given at the end."
},
{
  "role": "user",
  "content": [
    {"type": "text", "text": "example 1:"},
    {"type": "text", "text": "questionDirection: ... direction image"}, 
    {"type": "image_url", "image_url": {"url": "data:image/png;base64,..."}},
    {"type": "text", "text": "\nquestionText: ... question image"},
    {"type": "image_url", "image_url": {"url": "data:image/png;base64,..."}},
    {"type": "text", "text": "\nanswerChoices: \n 1. ... \n 2. ... \noptionImages: \n..."},
    {"type": "text", "text": "{'answer': <correct_option_number>, 'explanation': ...}"},
    {"type": "text", "text": "So the solution is ... solution image"},
    {"type": "image_url", "image_url": {"url": "data:image/png;base64,..."}},
    {"type": "text", "text": "\n\n..."},  
    {"type": "text", "text": "example 2:"},  
    ...
    {"type": "text", "text": "example 3:"},
    ...
    {"type": "text", "text": "\n\n..."}, 
    {"type": "text", "text": "now solve this question..."}
    {"type": "text", "text": "questionDirection: ..."}, 
    {"type": "image_url", "image_url": {"url": "data:image/png;base64,..."}},
    {"type": "text", "text": "\nquestionText: ..."},
    {"type": "image_url", "image_url": {"url": "data:image/png;base64,..."}},
    {"type": "text", "text": "\nanswerChoices: \n 1. ... \n 2. ... \noptionImages: \n..."},
    {"type": "text", "text": "Answer in the json format as follows: \n {'answer': <correct_option_number>, 'explanation': <explanation>}"} 
  ]
}
\end{lstlisting}


\subsubsection{Image Only}
\textbf{Zero shot}
\centering
\begin{lstlisting}[language=json, breaklines=true, caption={Zero-shot prompt template for image-only question answering. The prompt includes a system instruction to solve the problem and provide the answer in JSON format, followed by the input image encoded in base64.}]
{
  "role": "system",
  "content": "You are a brilliant problem solver... answer the correct option from the given choices along with a explanation. Answer in form of json in this format: {'answer': <correct_option_number>, 'explanation': <explanation>}"
},
{
  "role": "user",
  "content": [
    {"type": "image_url", "image_url": {"url": "data:image/png;base64,..."}}, 
  ]
}
\end{lstlisting}

\textbf{Few shot}
\centering
\begin{lstlisting}[language=json, breaklines=true, caption={Few-shot prompt template for image-only question answering. The template demonstrates two examples, each with a question image, the correct answer, an explanation, and potentially solution images. It then presents a new question image and asks for a structured answer in JSON format.}]
{
  "role": "system",
  "content": "Understand the following problems carefully ..."
},
{
  "role": "user",
  "content": [
    {"type": "text", "text": "You are a brilliant problem solver... Please select the correct answer choice."},
    {"type": "text", "text": "example 1:"},
    {"type": "image_url", "image_url": {"url": "data:image/png;base64,..."}}, 
    {"type": "text", "text": "\n so the solution to this example is as follows\n{'answer': <correct_option_number>, 'explanation': ...}"},
    {"type": "image_url", "image_url": {"url": "data:image/png;base64,..."}},
    {"type": "text", "text": "\n\n..."},
    {"type": "text", "text": "example 2:"},
    ...
    {"type": "text", "text": "example 3:"},
    ...
    {"type": "text", "text": "\n\n\n now answer the following question"},
    {"type": "image_url", "image_url": {"url": "data:image/png;base64,..."}}, 
  ]
}
\end{lstlisting}

\subsubsection{Standard VQA}
\textbf{Zero Shot}
\centering
\begin{lstlisting}[language=json, breaklines=true, caption={Zero-shot prompt template for Standard VQA on GPT-4o. The template instructs the model to solve a multiple-choice question with reference to an image. The expected output is a JSON object with the correct answer number and a corresponding explanation.}]
{
  "role": "system",
  "content": "You are a brilliant problem solver. Solve the given multiple choice question. Answer in the json format as follows: \n {'answer': <correct_option_number>, 'explanation': <explanation>}"
},
{
  "role": "user",
  "content": [
    {"type": "text", "text": "refer to this image for references in question: \n"},
    {"type": "image_url", "image_url": {"url": "data:image/png;base64,..."}},
    {"type": "text", "text": "\nquestion: ... \n answer the question with the correct option number and explanation in the json format as follows: \n {'answer': <correct_option_number>, 'explanation': <explanation>}"}
  ]
}
\end{lstlisting}
\textbf{Few shot}
\centering
\begin{lstlisting}[language=json, breaklines=true, caption={Few-shot prompt template for Standard VQA on GPT-4o. The template showcases a few-shot example with question, images, answers, explanations, and optional solution images, followed by a new question for the model to answer in JSON format.}] 
[
    {"type": "system", "content": "You are a brilliant problem solver... First understand the provided questions and then answer the new question given at the end."},
    {"type": "user", "content": [
        {"type": "text", "text": "example 1:"},
        {"type": "text", "text": "refer to this image for references in question: \n"},
        {"type": "image_url", "image_url": {"url": "data:image/png;base64,..."}}, 
        {"type": "text", "text": "\nquestion: ... \n answer the question with the correct option number and explanation in the json format as follows: \n {'answer': <correct_option_number>, 'explanation': <explanation>}"},
        {"type": "text", "text": "so the answer to this question is as follows: \n {'answer': <correct_option_number>, 'explanation': ...}"},
        {"type": "text", "text": "refer to this image for solution: \n"},
        {"type": "image_url", "image_url": {"url": "data:image/png;base64,..."}}, 
        {"type": "text", "text": "\n..."},
        {"type": "text", "text": "\n now as you have got the idea of the questions, let's answer the following question with a thorough explanation: \n"},
        {"type": "text", "text": "refer to this image for references in question: \n"},
        {"type": "image_url", "image_url": {"url": "data:image/png;base64,..."}},
        {"type": "text", "text": "\nquestion: ... \n answer the question with the correct option number and explanation in the json format as follows: \n {'answer': <correct_option_number>, 'explanation': <explanation>"}
    ]}
]
\end{lstlisting}

\section{Examples of the dataset} \label{Appendix:Examples}

\begin{flushleft}
\textbf{1. Series Problem}

Here we need to find missing element in number, alphabet or alpha-numeric series

\textbf{Question:} Find the missing element in the following series:

\[ 4, 6, 6, 15, 8, 28, 10, \underline{\hspace{1cm}} \]

\begin{enumerate}
    \item 36 
    \item 39 
    \item 45 
    \item 38 
\end{enumerate}

\textbf{Answer:} 3

\textbf{Explanation:} 
First series:    $ 4, 6, 8, 10 $ 

Second series : $ 6,15,28 $, ?  

Differences in the second series are 9, 13, 17 etc.  Hence the next term is $ 28+17=45 $.

\end{flushleft}
\begin{flushleft}
\textbf{2. Alphabet Test}

\noindent This question involves operations on the English alphabet:
\\
 
\noindent \textbf{Question:} Which letter should be the ninth letter to the left of the ninth letter from the right, if the first half of the given alphabet is reversed? 

\[ A \ B \ C \ D \ E \ F \ G \ H \ I \ J \ K \ L \ M \] 
\[ N \ O \ P \ Q \ R \ S \ T \ U \ V \ W \ X \ Y \ Z \]

\begin{enumerate}
    \item 1, 2, 3, 4, 5
    \item 1, 5, 3, 4, 2
    \item 5, 1, 2, 3, 4
    \item 3, 1, 5, 2, 4
\end{enumerate}

\textbf{Answer:} 2

\textbf{Explanation:} The new alphabet series is M LK J I H G F E D C B A N O P Q R S T U V W X Y Z. 

The 9th letter from right is $ R $ and the ninth letter to the left of $ R $ is $ E $.  
\end{flushleft}

\begin{flushleft}
    
\textbf{3. Classification/Odd One Out}

\noindent This category of questions requires identifying the option that doesn't belong with the others.

\noindent \textbf{Question:} Find the odd term/wrong term or which is different from the rest three terms.

\begin{enumerate}
\item 31:96
\item 15:63
\item 22:91
\item 23:95
\end{enumerate}

\textbf{Answer:} 1

\textbf{Explanation:} The pattern is: first number ×4+3= second number.

\[
\begin{aligned}
22 \times 4 + 3 &= 91 \\
15 \times 4 + 3 &= 63 \\
23 \times 4 + 3 &= 95 \\
31 \times 4 + 3 &= 127
\end{aligned}
\]

The first option (31:96) does not follow this pattern.
\end{flushleft}

\begin{flushleft}
\textbf{4. Analogy Problems}

\noindent This category of questions presents an analogy where the first two terms have a relationship. You need to identify the pair that shares the same relationship.

\noindent \textbf{Question:} Square : Cube :: 

\begin{enumerate}
    \item Rectangle : Cuboid 
    \item Triangle : Square
    \item Quadrilateral : Cuboid 
    \item Cuboid : Rectangle 
\end{enumerate}

\textbf{Answer:} 1

\textbf{Explanation:} A cube is the three-dimensional extension of a square. Similarly, a cuboid is the three-dimensional extension of a rectangle.
\end{flushleft}

\begin{flushleft}
\textbf{ 5. Coding-Decoding}

\noindent In this category, an example of a code is given. You need to infer the rule and then code/decode a new example.

\noindent \textbf{Question:} If A = 2, T = 40, and ACT = 48, then TAKE = ?

\begin{enumerate}
\item 68
\item 58
\item 74
\item 76
\end{enumerate}

\textbf{Answer:} 3

\textbf{Explanation:}

The rule is: (Position value of the letter in the alphabet) ×2 = Code

\[
\begin{aligned}
A &= 1 \times 2 = 2 \\
T &= 20 \times 2 = 40 \\
ACT &= (1 + 3 + 20) \times 2 = 48 \\
TAKE &= (20 + 1 + 11 + 5) \times 2 = 74
\end{aligned}
\]
\end{flushleft}

\begin{flushleft}
\textbf{6. Number and Ranking Problems}

\noindent This category of questions involves arranging items in a logical sequence.

\noindent \textbf{Question:} Arrange the following words in a meaningful sequence.

1. Key, 
2. Door, 
3. Lock, 
4. Room, 
5. Switch on
\begin{enumerate}
\item 5, 1, 2, 4, 3
\item 4, 2, 1, 5, 3
\item 1, 3, 2, 4, 5
\item 1, 2, 3, 5, 4
\end{enumerate}

\textbf{Answer:} 3

\textbf{Explanation:} The logical order of actions is:

You need the Key (1)
To open the Lock (3)
On the Door (2)
To enter the Room (4)
And then Switch on (5) the lights.
\end{flushleft}

\begin{flushleft}
\textbf{7. Blood Relation}

\noindent This category involves decoding relationships based on given symbols and then analyzing statements to determine their accuracy.

\noindent \textbf{Question:} Which of the following is correct?

\noindent Symbols and their meanings:

* $P = Q$: $Q$ is the father of $P$
* $P * Q$: $P$ is the sister of $Q$
* $P ? Q$: $Q$ is the mother of $P$
* $P \ Q$: $P$ is the brother of $Q$
* $P \subset Q$: $Q$ is the son of $P$
* $P \times Q$: $P$ is the daughter of $Q$

\begin{enumerate}
    \item $V \times T * P$ means $P$ is the maternal uncle of $V$.
    \item $D ? V \times T$ means $D$ is the granddaughter of $T$.
    \item $L \subset M \ R$ means $R$ is the paternal uncle of $L$.
    \item $M \ R * R \ D ? V$ means $M$ is the son of $V$.
\end{enumerate}

\textbf{Answer:} 4

\textbf{Explanation:} 

Let's break down each statement:

1. $V \times T * P$: 
    * $V \times T$: $V$ is the daughter of $T$
    * $T * P$: $T$ is the sister of $P$
    * Conclusion: $P$ could be the maternal uncle *or* maternal aunt of $V$. So, this statement is incorrect.

2. $D ? V \times T$:
    * $D ? V$: $V$ is the mother of $D$
    * $V \times T$: $V$ is the daughter of $T$
    * Conclusion: $D$ could be the grandson *or* granddaughter of $T$. So this statement is incorrect

3. $L \subset M \ R$:
    * $L \subset M$: $M$ is the son of $L$
    * $M \ R$: $M$ is the brother of $R$
    * Conclusion: $R$ is the son of $L$. So, $R$ is the paternal uncle of $M$, not $L$. This statement is incorrect

4. $M \ R * R \ D ? V$:
    * $M \ R$: $M$ is the brother of $R$
    * $R * D$: $R$ is the sister of $D$
    * $D ? V$: $V$ is the mother of $D$
    * Conclusion: $M$ is the brother of $R$, who is the daughter of $V$. This means $M$ is the son of $V$. This statement is correct. 

Therefore, the correct statement is option 4. 
\end{flushleft}

\begin{flushleft}
    \textbf{8. Mathematical Operations}

\noindent This category involves deducing the underlying mathematical operation from given examples and applying it to a new problem.

\noindent \textbf{Question:} If $37 + 42 = 16$, $43 + 54 = 16$, and $25 + 34 = 14$, then $65 + 35 = ?$

\begin{enumerate}
    \item 100
    \item 91
    \item 18
    \item 19 
\end{enumerate}

\textbf{Answer:} 4

\textbf{Explanation:} 

The operation is to sum the individual digits of the numbers being 'added'.

\[
\begin{aligned}
37 + 42 &\rightarrow 3 + 7 + 4 + 2 = 16 \\
43 + 54 &\rightarrow 4 + 3 + 5 + 4 = 16 \\
25 + 34 &\rightarrow 2 + 5 + 3 + 4 = 14 \\
65 + 35 &\Rightarrow 6 + 5 + 3 + 5 = 19
\end{aligned}
\]
\end{flushleft}

\begin{flushleft}
    \textbf{9. Direction Sense}

\noindent This category involves understanding directions and calculating distances based on movements.

\noindent \textbf{Question:} A man walks 1 km towards East and then he turns to South and walks 5 km. Again he turns to East and walks 2 km, after this he turns to North and walks 9 km. Now, how far is he from his starting point?

\begin{enumerate}
    \item 3 km
    \item 4 km
    \item 5 km
    \item 7 km 
\end{enumerate}

\textbf{Answer:} 3

\textbf{Explanation:} 



* A: Starting point\\
* B: 1 km East of A\\
* C: 5 km South of B\\
* D: 2 km East of C\\
* E: 9 km North of D

We want to find the distance AE.

* DF = BC = 5 km \\
* EF = (DE - DF) = (9 - 5) km = 4 km\\
* BF = CD = 2 km\\
* AF = AB + BF = 1 + 2 = 3 km
Using the Pythagorean theorem on triangle AEF:

\[ \begin{aligned}
AE &= \sqrt{AF^2 + EF^2} \\
   &= \sqrt{3^2 + 4^2} = \sqrt{25} = 5 \text{ km} 
\end{aligned} \] 
Therefore, the man is 5 km from his starting point.

\end{flushleft}

\begin{flushleft}
\textbf{10. Venn Diagrams}

\noindent This category involves interpreting information represented in Venn diagrams.

\noindent \textbf{Directions:} In the following diagram, three classes of population are represented by three figures.

The triangle represents school teachers.
The square represents married persons.
The circle represents persons living in joint families.

\noindent \textbf{Question:} School teachers who are married but do not live in joint families are represented by

\textbf{Question Image}
\begin{figure}[h]
    \centering
    \includegraphics[width=0.5\linewidth]{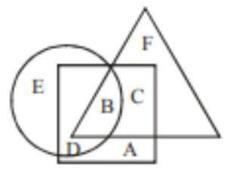}
    \label{fig:venn_diagram_problem}
    \caption{Venn Diagram Problem: Question 10}
\end{figure}
\begin{enumerate}
\item C
\item F
\item A
\item D
\end{enumerate}

\textbf{Answer:} 1

\textbf{Explanation:}

Married teachers are represented by the intersection of the triangle (teachers) and the square (married), which includes regions B and C.
We need those who do not live in joint families, so we exclude the circle.
Only region C satisfies both conditions: married teachers outside the joint family circle.
Therefore, the answer is C.
\end{flushleft}

\begin{flushleft}
    \textbf{11. Time and Clock}

\noindent This category involves calculations related to days, dates, and calendars. 

\noindent \textbf{Question:} If it was Saturday on 17th December, 2002, what was the day on 22nd December, 2004?

\begin{enumerate}
    \item Monday
    \item Tuesday
    \item Wednesday
    \item Sunday 
\end{enumerate}

\textbf{Answer:} 4

\textbf{Explanation:} 

The period from 17th Dec. 2002 to 16th Dec. 2003 is 365 days (52 weeks + 1 day).  \\
So, 16th Dec. 2003 is also a Saturday. \\
The period from 16th Dec. 2003 to 15th Dec. 2004 is 366 days (2004 is a leap year). \\
So, 15th Dec. 2004 is also a Saturday. \\
Counting forward, 22nd Dec. 2004 is a Sunday.

Therefore, the answer is Sunday. 

\end{flushleft}

\begin{flushleft}
\textbf{12. Missing Character}

\noindent This category involves predicting a missing element within a figure, often requiring spatial reasoning.

\noindent \textbf{Question:} Find the missing term in the following figure:
\begin{figure}[h]
    \centering
    \includegraphics[width=0.5\linewidth]{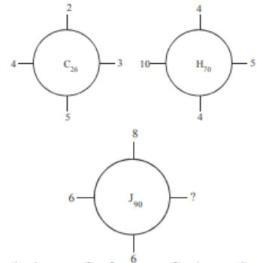}
    \label{fig:missing_character_problem}
    \caption{Missing Character Problem: Question 12}
\end{figure}
\begin{enumerate}
    \item 1 
    \item 2 
    \item 4 
    \item 10 
\end{enumerate}

\textbf{Answer:} 3

\textbf{Explanation:} 

The number inside the circle is obtained by the following rule:

* Sum the upper number, the lower number, and the alphabetical position of the letter.
* Multiply this sum by the number on the right.
* Subtract the number on the left from the product.

Applying this rule to the given examples:

* $(2 + C + 5) \times 3 - 4 = (2 + 3 + 5) \times 3 - 4 = 26$
* $(4 + H + 4) \times 5 - 10 = (4 + 8 + 4) \times 5 - 10 = 70$

Let the missing number be  'x'. Then,

* $(8 + J + 6) \times x - 6 = 90$
* $(8 + 10 + 6) \times x = 96$
* $x = 4$

Therefore, the missing number is 4.
\end{flushleft}

\begin{flushleft}
\textbf{13. Non-Verbal Reasoning - Series Continuation}

\noindent \textbf{Directions:} Each question consists of five problem figures (A, B, C, D, and E) followed by four answer figures (1, 2, 3, and 4). Select the figure that continues the series established by the problem figures.

\noindent \textbf{Problem Figures:}

\begin{figure}[h]
    \centering
    \includegraphics[width=0.8\linewidth]{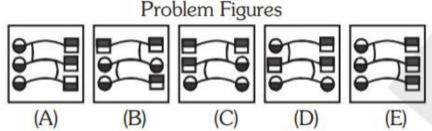}
    \caption{Non-Verbal Reasoning -Series Continuation: Question 13}
        \label{fig:non_verbal_reasoning_problem:0}
\end{figure}

\begin{enumerate}

    \item \begin{minipage}[c]{0.35\linewidth} 
               \includegraphics[width=\linewidth]{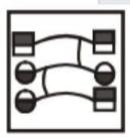}
           \end{minipage}\hfill
    \item \begin{minipage}[c]{0.35\linewidth}
               \includegraphics[width=\linewidth]{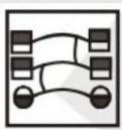}
           \end{minipage}
    \item \begin{minipage}[c]{0.35\linewidth}
               \includegraphics[width=\linewidth]{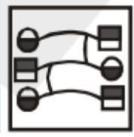}
           \end{minipage}\hfill
    \item \begin{minipage}[c]{0.35\linewidth}
               \includegraphics[width=\linewidth]{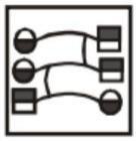}
           \end{minipage}
\end{enumerate}

\textbf{Answer:} 1

\textbf{Explanation:} 
The following figures explain the pattern.
\begin{figure}[h]
    \centering
    \includegraphics[width=0.9\linewidth]{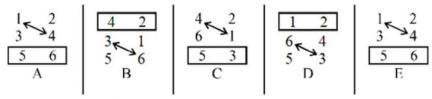}
    \caption{Problem 13 Solution Explanation.}
    \label{fig:non_verbal_reasoning_problem:1}
\end{figure}
\begin{figure}[h]
    \centering
    \includegraphics[width=0.5\linewidth]{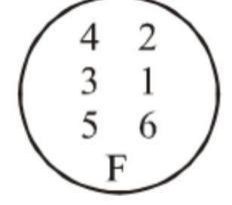}
    \caption{Problem 13 Solution.}
    \label{fig:non_verbal_reasoning_problem:2}
\end{figure}

\end{flushleft}

\begin{flushleft}
    \textbf{14. Non-Verbal Classification/Odd One Out}
\noindent \textbf{Questions:} Which is the odd one out ?   
\begin{figure}[!h]
    \centering
    \includegraphics[width=1\linewidth]{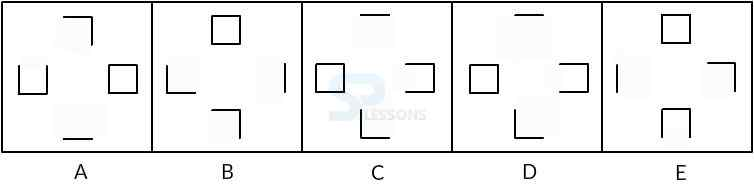}
    \caption{Problem 14}
    \label{fig:non_verbal_reasoning_problem:2}
\end{figure}

\textbf{Answer:} B

\textbf{Explanation:} 
Each one of the figures except figure B, contains - one complete square, one cup-shaped element having three sides, one 'L'-shaped element having two sides and one straight line. Therefore, the figure B is different from the rest.
\end{flushleft}

\begin{flushleft}
\textbf{15. Non-Verbal Reasoning - Analogy}

\noindent \textbf{Directions:} Each question consists of two sets of figures: A, B, C, and D. A definite relationship exists between figures A and B. Establish a similar relationship between figures C and D by choosing a suitable figure D from the answer set. 

\begin{figure}[h]
    \centering
    \includegraphics[width=0.9\linewidth]{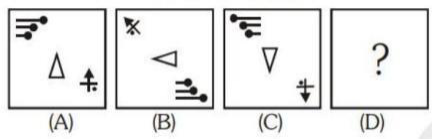}
    \caption{Problem Non-Verbal Reasoning - Analogy.}
    \label{fig:AP32_problem_0}
\end{figure}
\noindent \textbf{Answer Figures:}

\begin{enumerate}
    \item \begin{minipage}[c]{0.45\linewidth}
               \includegraphics[width=\linewidth]{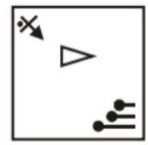}
           \end{minipage}\hfill
    \item \begin{minipage}[c]{0.45\linewidth}
               \includegraphics[width=\linewidth]{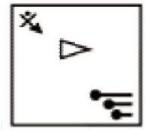}
           \end{minipage}
    \item \begin{minipage}[c]{0.45\linewidth}
               \includegraphics[width=\linewidth]{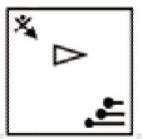}
           \end{minipage}\hfill
    \item \begin{minipage}[c]{0.45\linewidth}
               \includegraphics[width=\linewidth]{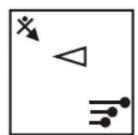}
           \end{minipage}
\end{enumerate}

\textbf{Answer:} 3

\textbf{Explanation:} By observation. 
\end{flushleft}

\begin{flushleft}

\textbf{16. Incomplete Figure}

\noindent This category involves identifying the missing part of a figure to complete it.

\noindent \textbf{Question:} Which of the answer figures will complete the matrix figure?

\noindent \textbf{Question Figure:}

\begin{figure}[h]
    \centering
    \includegraphics[width=0.7\linewidth]{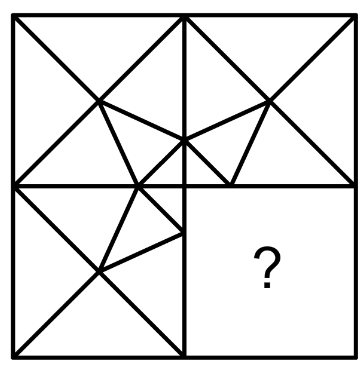}
    \caption{Question 16 Problem}
    \label{fig:incomplete_figure_problem}
\end{figure}

\noindent \textbf{Answer Figures:}

\begin{enumerate}
    \item \begin{minipage}[c]{0.35\linewidth}
               \includegraphics[width=\linewidth]{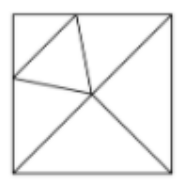}
           \end{minipage}\hfill
    \item \begin{minipage}[c]{0.35\linewidth}
               \includegraphics[width=\linewidth]{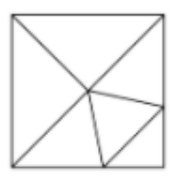}
           \end{minipage}
    \item \begin{minipage}[c]{0.35\linewidth}
               \includegraphics[width=\linewidth]{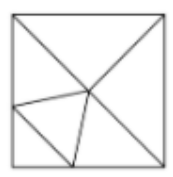}
           \end{minipage}\hfill
    \item \begin{minipage}[c]{0.35\linewidth}
               \includegraphics[width=\linewidth]{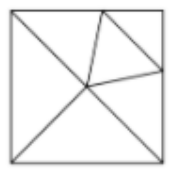}
           \end{minipage}
\end{enumerate}

\textbf{Answer:} 1

\textbf{Explanation:} By observation. 
\end{flushleft}

\begin{flushleft}
    \textbf{17. Mirror Image}

\noindent This category involves identifying the mirror image of a given figure.

\noindent \textbf{Question:} The mirror image of the given diagram is:

\noindent \textbf{Question Figure:}
\begin{figure}[h]
    \centering
    \includegraphics[width=0.5\linewidth]{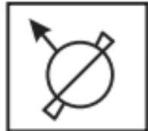}
    \caption{17.1}
    \label{fig:mirror_image_problem}
\end{figure}

\noindent \textbf{Answer Figures:}

\begin{enumerate}
    \item \begin{minipage}[c]{0.45\linewidth}
               \includegraphics[width=\linewidth]{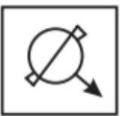}
           \end{minipage}\hfill
    \item \begin{minipage}[c]{0.45\linewidth}
               \includegraphics[width=\linewidth]{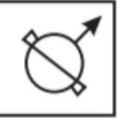}
           \end{minipage}
    \item \begin{minipage}[c]{0.45\linewidth}
               \includegraphics[width=\linewidth]{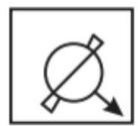}
           \end{minipage}\hfill
    \item \begin{minipage}[c]{0.45\linewidth}
               \includegraphics[width=\linewidth]{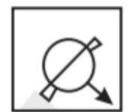}
           \end{minipage}
\end{enumerate}

\textbf{Answer:} 2

\textbf{Explanation:} By observation.
\end{flushleft}

\begin{flushleft}
    \textbf{18. Cube and Dice}

\noindent This category involves visualizing the folding of a 2D net into a 3D cube and identifying possible resulting cubes.

\noindent \textbf{Question:} Select from the alternatives, the box(es) that can be formed by folding the sheet shown in the figure. 

\noindent \textbf{Question Figure:}

\begin{center}
\includegraphics[width=0.4\linewidth]{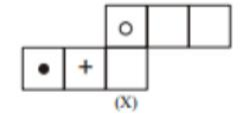}
\includegraphics[width=0.4\linewidth]{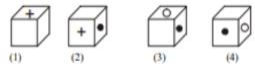}
\end{center}

\begin{enumerate}
    \item 1 only 
    \item 1, 2, and 3 only 
    \item 2 and 3 only 
    \item 1, 2, 3, and 4 
\end{enumerate}

\textbf{Answer:} 4

\textbf{Explanation:} 

When the sheet is folded to form a cube:

* The face with a dot will be opposite a blank face.
* The face with a "+" sign will be opposite another blank face
* The face with a circle will be opposite the third blank face

Considering these relationships, all four cubes shown in the options (1, 2, 3, and 4) can be formed. 

\end{flushleft}

\begin{flushleft}
\textbf{19. Paper Folding and Cutting}

\noindent \textbf{Directions:} In the following questions, a square sheet of paper is folded along the dotted lines, and then cuts are made on it. Select the figure from the given choices that shows how the sheet would look when opened.


    \begin{center}
        
    \includegraphics[width=0.55\linewidth]{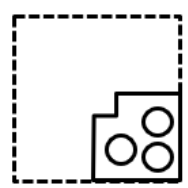}
    \end{center}

\noindent Choose the correct figure.

\begin{enumerate}
    \item \begin{minipage}[c]{0.45\linewidth}
               \includegraphics[width=\linewidth]{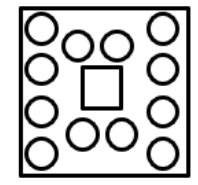}
           \end{minipage}\hfill
    \item \begin{minipage}[c]{0.45\linewidth}
               \includegraphics[width=\linewidth]{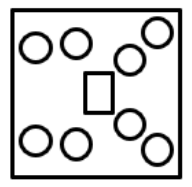}
           \end{minipage}
    \item \begin{minipage}[c]{0.45\linewidth}
               \includegraphics[width=\linewidth]{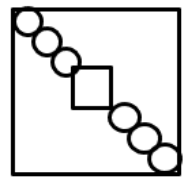}
           \end{minipage}\hfill
        \item \begin{minipage}[c]{0.45\linewidth}
               \includegraphics[width=\linewidth]{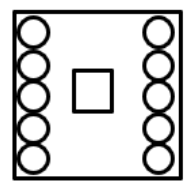}
           \end{minipage}
\end{enumerate}

\textbf{Answer:} 1

\textbf{Explanation:} By observation.
\end{flushleft}

\begin{flushleft}
\textbf{20. Embedded Figure}

\noindent \textbf{Directions:} In the following question, there is a question figure, which is embedded in one of the answer figures. Trace out the correct figure.

\noindent \textbf{Question Figure:}
\begin{center}
    \includegraphics[width=0.5\linewidth]{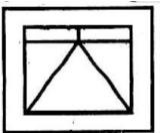}

\end{center}

\noindent \textbf{Answer Figures:}

\begin{enumerate}
    \item \begin{minipage}[c]{0.35\linewidth}
               \includegraphics[width=\linewidth]{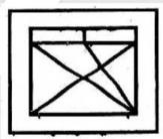}
           \end{minipage}\hfill
    \item \begin{minipage}[c]{0.35\linewidth}
               \includegraphics[width=\linewidth]{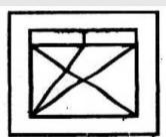}
           \end{minipage}
    \item \begin{minipage}[c]{0.35\linewidth}
               \includegraphics[width=\linewidth]{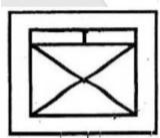}
           \end{minipage}\hfill
    \item \begin{minipage}[c]{0.35\linewidth}
               \includegraphics[width=\linewidth]{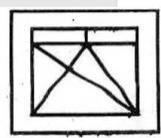}
           \end{minipage}
\end{enumerate}

\textbf{Answer:} 4

\end{flushleft}

\begin{flushleft}
    \section*{21. Puzzle Test}

\noindent This category involves solving general puzzles based on provided information.

\noindent \textbf{Directions:} Read the following information carefully and answer the question that follows:

\begin{enumerate}[label=(\roman*)]
    \item There is a group of five persons: A, B, C, D, and E.
    \item One of them is a Teacher, one is a Doctor, one is a Journalist, one is an Industrialist, and one is an Advocate.
    \item Three of them - A, C, and the Advocate - prefer tea to coffee.
    \item Two of them - B and the Journalist - prefer coffee to tea.
    \item The Industrialist, D, and A are friends, but two of these prefer coffee to tea.
    \item The Teacher is C's brother.
\end{enumerate}

\noindent \textbf{Question:} Who is the Teacher?

\begin{enumerate}
    \item B
    \item A
    \item C
    \item D 
\end{enumerate}

\textbf{Answer:} 2

\textbf{Explanation:} 

From the given information, we can deduce:

* A (Teacher), C (Doctor), E (Advocate) prefer tea to coffee.
* B (Industrialist), D (Journalist) prefer coffee to tea.

\end{flushleft}

\begin{flushleft}
\textbf{22. Figure Partition}

\noindent This category involves counting specific shapes or components within a given figure.

\noindent \textbf{Question:} The number of triangles in the following figure is:

\begin{center}
\includegraphics[width=0.5\linewidth]{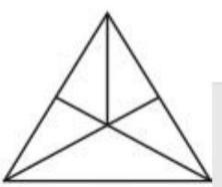} 
\end{center}

\begin{enumerate}
    \item 9
    \item 10
    \item 11
    \item 12 
\end{enumerate}

\textbf{Answer:} 4

\textbf{Explanation:} 

By careful observation of the figure, we can count a total of 12 triangles.

\end{flushleft}

\begin{flushleft}
    \section*{23. Dot Problem}

\noindent \textbf{Directions:} Select the alternative which satisfies the same conditions of placement of dots as shown in the figure. 

\noindent \textbf{Question Figure:}

\begin{center}
\includegraphics[width=0.4\linewidth]{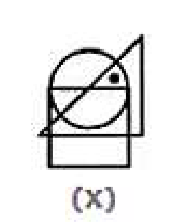} 
\end{center}

\noindent \textbf{Answer Figures:}


\begin{figure}[!h]
    \centering
    \includegraphics[width=1\linewidth]{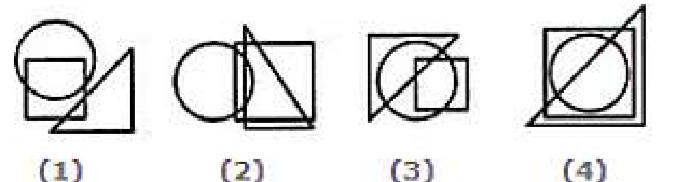}
    \label{fig:enter-label}
\end{figure}

\textbf{Answer:} 3 \\
\textbf{Explanation:} In figure (X), the dot is placed in the region which is common to the circle and triangle. Now, we have to search similar common region in the four options. Only in figure (3), we find such a region which is common to the circle and triangle.
\end{flushleft}

\begin{flushleft}
    \textbf{24. Syllogisms}

\noindent This category involves evaluating logical arguments and identifying valid conclusions based on given statements.

\noindent \textbf{Directions:} Each question consists of five or six statements followed by options containing three statements in a specific order. Choose the option which indicates a valid argument, where the third statement is a conclusion drawn from the preceding two statements.

\noindent \textbf{Statements:}

\begin{enumerate}[label=\Alph*.]
    \item All synopses are poets.
    \item Some synopses are mentors.
    \item Some X are not mentors.
    \item All X are poets.
    \item All synopses are mentors
    \item All synopses are X.
\end{enumerate}

\noindent \textbf{Options:}

\begin{enumerate}
    \item ABC 
    \item AEC 
    \item FEC 
    \item DFA 
\end{enumerate}

\textbf{Answer:} 4

\textbf{Explanation:} 

Let's analyze each option:

\begin{enumerate}
    \item \textbf{ABC}: Irrelevant 
    \item \textbf{AEC}: Irrelevant 
    \item \textbf{FEC}: The conclusion may or may not be true.
    \item \textbf{DFA}: 
        * D: All X are poets
        * F: All synopses are X
        * Conclusion: All synopses are poets. This is a valid conclusion as it follows from the first two statements
\end{enumerate}

Thus, the correct answer is option 4.

\end{flushleft}

\begin{flushleft}

\textbf{25. Statement \& Conclusions}

\noindent This category involves making inferences based on given statements.

\noindent \textbf{Directions:} In the question below, two statements are given followed by two conclusions (I and II). Take the statements to be true and then decide which of the conclusions logically follows.

\noindent \textbf{Statements:} The average number of students per teacher is 50 in the urban area, whereas it is 60 in rural areas. The national average is 55.

\noindent \textbf{Conclusions:}

\begin{enumerate}[label=\Roman*.]
    \item The student-teacher ratio in the rural areas is higher than in the urban areas.
    \item More students study with the same teacher in the rural areas as compared to those in the urban areas.
\end{enumerate}

\noindent \textbf{Options:}

\begin{enumerate}
    \item if conclusion I follows
    \item if conclusion II follows
    \item if either conclusion I or II is implicit
    \item if neither conclusion I nor II follows
\end{enumerate}

\textbf{Answer:} 2

\textbf{Explanation:} 

* Without absolute figures (total number of students and teachers), we cannot conclude anything about the student-teacher ratio (Conclusion I).

* The average number of students per teacher is higher in rural areas (60) compared to urban areas (50). This directly implies that more students study with the same teacher in rural areas (Conclusion II).

Therefore, only conclusion II follows. 

\end{flushleft}

\begin{flushleft}
    \textbf{26. Data Sufficiency}

\noindent This category involves determining whether given statements provide enough information to answer a question.

\noindent \textbf{Directions:} Each question has a problem and two statements (I and II). Decide if the information in the statements is sufficient for answering the problem.

\noindent \textbf{Question:} Who is the father of M?

\noindent \textbf{Statements:}

\begin{enumerate}[label=\Roman*.]
    \item A and B are brothers.
    \item B's wife is the sister of M's wife.
\end{enumerate}

\noindent \textbf{Options:}

\begin{enumerate}
    \item if the data in statement I alone are sufficient to answer the question
    \item if the data in statement II alone are sufficient to answer the question
    \item if the data either in I or II alone are sufficient to answer the question
    \item if the data even in both the statements together are not sufficient to answer the question
    \item if the data in both the statements together are needed
\end{enumerate}

\textbf{Answer:} 4

\textbf{Explanation:} 

* From statement II, we conclude that B is the brother-in-law of M.
* Even combining both statements, we cannot determine who the father of M is.

Therefore, the data in both statements together are not sufficient to answer the question. 

\end{flushleft}

\end{document}